%% file: main.tex
\documentclass[runningheads]{llncs}
\usepackage{graphicx}

\usepackage{tikz}
\usepackage{comment}
\usepackage{amsmath,amssymb} 
\usepackage{color}

\usepackage[accsupp]{axessibility}  

\usepackage{bm}
\usepackage{bbm}
\usepackage{multirow}
\usepackage{xcolor}
\usepackage{xspace}
\usepackage{booktabs}
\usepackage{style}
\usepackage{caption}
\usepackage{subcaption}
\usepackage{hyperref}
\usepackage{soul}
\usepackage[page]{appendix}
\usepackage{orcidlink}

\begin{document}
\pagestyle{headings}
\mainmatter
\def\ECCVSubNumber{7}  

\title{Future Object Detection with Spatiotemporal Transformers} 

\titlerunning{Future Object Detection with Spatiotemporal Transformers}
%
\author{Adam Tonderski\inst{1,2}\orcidlink{0000-0002-2160-4386} \and
Joakim Johnander\inst{1,3}\orcidlink{1111-2222-3333-4444} \and
Christoffer Petersson\inst{1,4}\orcidlink{0000-0002-9203-558X} \and
Kalle Åström\inst{2}\orcidlink{0000-0002-8689-7810}
}
\authorrunning{A. Tonderski et al.}
%
\institute{Zenseact AB, Sweden \\
\email{\{firstname.lastname\}@zenseact.com} \and
Lund University, Sweden \\
\email{\{adam.tonderski,karl.astrom\}@math.lth.se} \and
Department of Electrical Engineering, Linköping University, Sweden \and
Chalmers University of Technology, Sweden \\
}

\maketitle

\input{inputs/abstract}
\input{inputs/introduction}
\input{inputs/related_work}
\input{inputs/method}
\input{inputs/experiments}
\input{inputs/conclusion}

\input{inputs/acknowledgements}

\clearpage
%
%
\bibliographystyle{splncs04}
\bibliography{references}

\input{inputs/supplementary}

\end{document}

%% file: inputs/abstract.tex
\begin{abstract}

We propose the task Future Object Detection, in which the goal is to predict the bounding boxes for all visible objects in a future video frame. While this task involves recognizing temporal and kinematic patterns, in addition to the semantic and geometric ones, it only requires annotations in the standard form for individual, single (future) frames, in contrast to expensive full sequence annotations. We propose to tackle this task with an end-to-end method, in which a detection transformer is trained to directly output the future objects. In order to make accurate predictions about the future, it is necessary to capture the dynamics in the scene, both object motion and the movement of the ego-camera. To this end, we extend existing detection transformers in two ways. First, we experiment with three different mechanisms that enable the network to spatiotemporally process multiple frames. Second, we provide ego-motion information to the model in a learnable manner. We show that both of these extensions  improve the future object detection performance substantially. Our final approach learns to capture the dynamics and makes predictions on par with an oracle for prediction horizons up to 100 ms, and outperforms all baselines for longer prediction horizons. By visualizing the attention maps, we observe that a form of tracking emerges within the network. Code is available at \rurl{github.com/atonderski/future-object-detection}.

\end{abstract}

%% file: inputs/introduction.tex
\section{Introduction}
Autonomous robots, such as self-driving vehicles, need to make predictions about the future in order to drive safely and efficiently.
Several forecasting tasks have been presented in the literature, such as predicting future RGB-values~\cite{mathieu2016deep,ranzato2014video,srivastava2015unsupervised,kalchbrenner2017video}; instance segmentations~\cite{luc2018predicting}; birds-eye view instance segmentations~\cite{hu2021fiery}; and trajectories of seen objects~\cite{styles2020multiple}. The latter three tasks involve making predictions about dynamic objects, such as humans, vehicles, or animals. These tasks require the ability to estimate the dynamics of the scene -- a challenging problem in itself -- and use that for extrapolation~\cite{wu2020future,luc2018predicting,styles2020multiple}. Furthermore, the tasks are inherently ambiguous and methods need to present multiple hypotheses.

\begin{figure*}[t]
     \centering
     \begin{subfigure}[b]{0.32\textwidth}
         \centering
         \includegraphics[width=\textwidth, trim=80 50 120 90, clip]{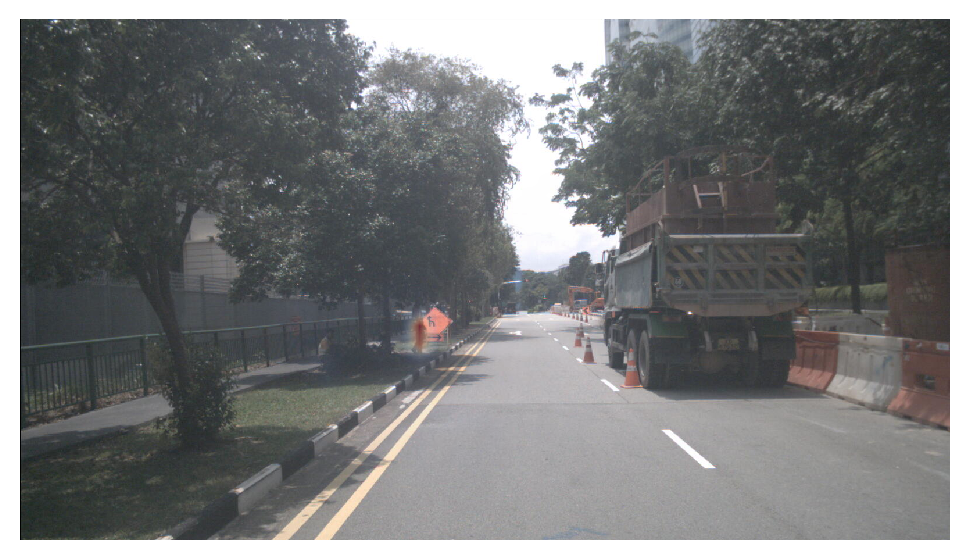}
        \caption*{$t=T-1$}
     \end{subfigure}
     \hfill
     \begin{subfigure}[b]{0.32\textwidth}
         \centering
         \includegraphics[width=\textwidth, trim=80 50 120 90, clip]{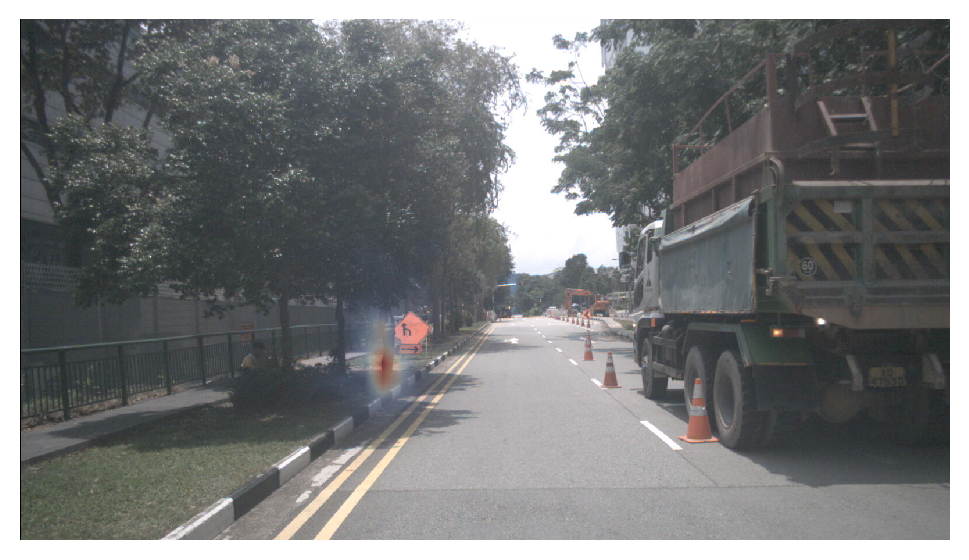}
         \caption*{$t=T$}
     \end{subfigure}
     \hfill
     \begin{subfigure}[b]{0.32\textwidth}
         \centering
         \includegraphics[width=\textwidth, trim=80 50 120 90, clip]{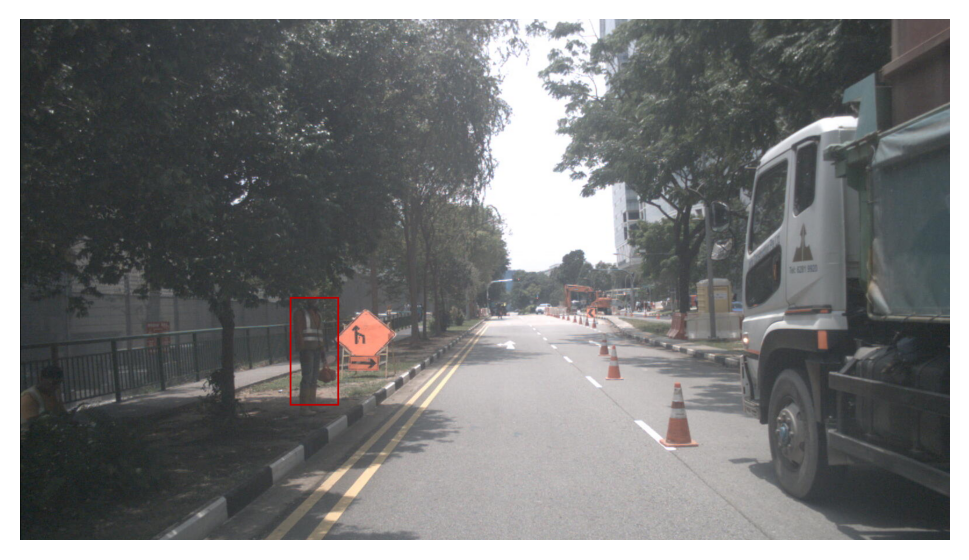}
         \caption*{$t=T+\tau$}
     \end{subfigure}
     \vspace{-3mm}
    \caption{
    We design a novel method for future object detection that takes a pair of frames, from integer timesteps $T-1$ (left) and $T$ (middle), as input, and outputs a prediction of the objects at a future timestep $T+\tau$ (right). During training, the model is only provided ground truth for the \emph{single} \emph{unseen} future frame $T+\tau$. In the left and middle figures, we show the attention maps associated to the predicted object at $T+\tau$. Our approach accurately predicts the pedestrian at the future timestep by attending to him in the previous (input) frames, indicating an emergent form of implicit object tracking.}
    \label{fig:cool-future-pred}
    \vspace{-6mm}
\end{figure*}

In this work, we develop and investigate methods aimed at predicting object locations in a future video frame. In contrast to the previously mentioned forecasting tasks, which rely on more complex annotations involving for example 3d bounding boxes, instance segmentations, or object trajectories, we extend traditional object detection and propose a new task -- Future Object Detection. The goal is to predict all visible objects in a future image, represented in exactly the same way as in traditional object detection, \ie by a set of confidence, class, and 2d bounding box triplets. A key benefit of this task is that a larger and more diverse dataset can be constructed as only standard object detection annotations for a single frame are required, for both training and evaluation. Nevertheless, we find that models learn to capture the dynamics of the scene and make accurate predictions of objects in a future frame. For example, we find that a form of object tracking emerges inside a model when it is trained for future object detection, see Figure~\ref{fig:cool-future-pred}. Another benefit of using the standard object detection output representation is that metrics are compatible and common methods such as DETR~\cite{carion2020end} and FasterRCNN~\cite{ren2015faster} can be readily adapted.

\begin{figure*}[t]
    \centering
    \includegraphics[width=0.95\linewidth, trim=0mm 10mm 0mm 2mm, clip]{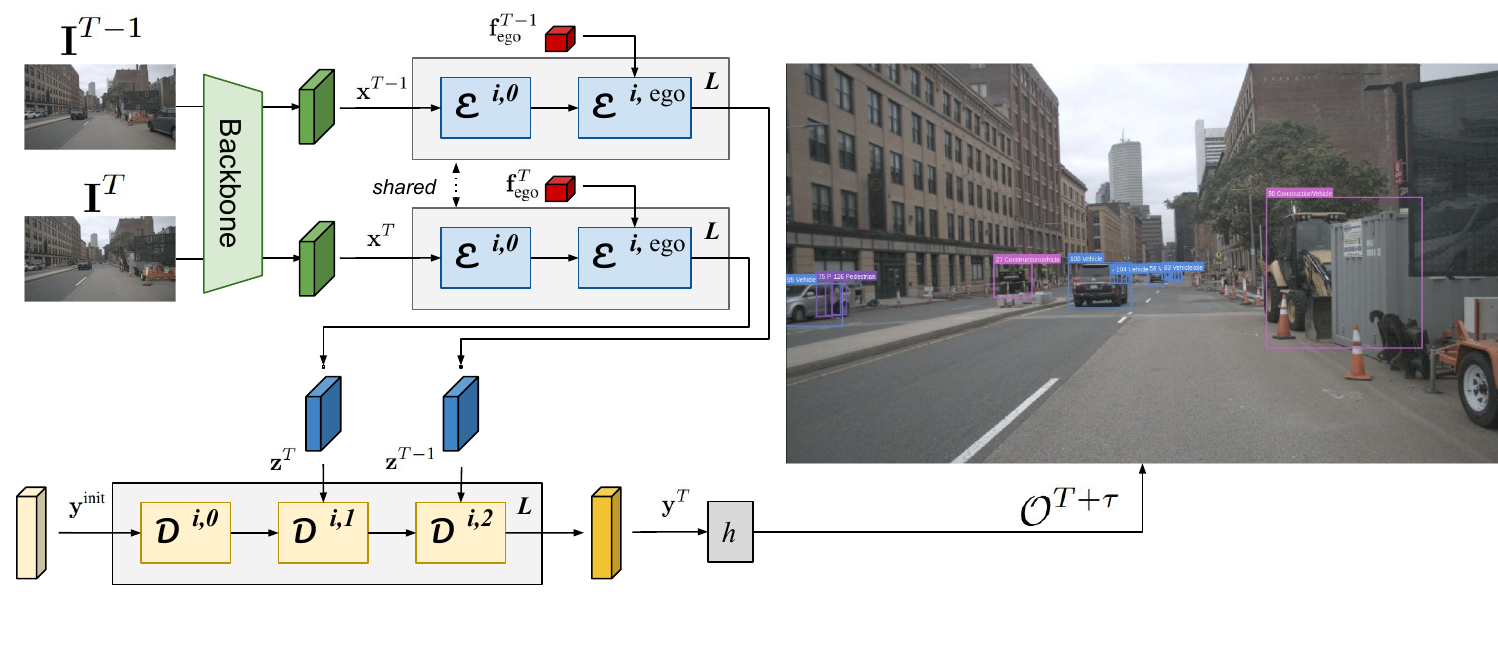}
    \vspace{-1mm}
    \caption{An overview of the proposed spatiotemporal transformer for future object detection, applied to two input frames. A backbone encodes each image. Each image is then fed into a transformer encoder $\enc$, in which ego-motion information is also incorporated. Next, a transformer decoder $\dec$ lets a set of queries $\mathbf{y}^\text{init}\in\mathbb{R}^{M\times D}$ gather information from the encoded feature maps through sequential cross-attention ($\dec^{i,1}, \dec^{i,2}$). Last, a small neural network head predicts the set of objects, $\mathcal{O}^{T+\tau}$, present in the future frame $T+\tau$.}
    \label{fig:overview}
    \vspace{-4mm}
\end{figure*}

A traditional way to tackle future object detection would be to divide it into several steps: (i) detect objects in every frame in a video sequence; (ii) associate detections across time and form tracks; (iii) extrapolate the tracks; and (iv) anticipate new objects that might appear. None of these steps are explicit in our approach. Instead, we propose to train a neural network for this problem in an end-to-end manner, using the formulation proposed by Carion \etal~\cite{carion2020end}. The neural network architectures proposed in their work and in subsequent works make use of a key mechanism: cross-attention. This enables an instance-level representation, originally referred to as object queries~\cite{carion2020end}, to gather information from the deep feature maps extracted from the image.\footnote{See Greff \etal~\cite{greff2020binding} for a discourse on different methods of representation.} This mechanism, due to its flexibility, provides a natural way to incorporate temporal information, in the form of past images, to enable the neural network to better anticipate the motion of the surrounding dynamic objects. Further, we learn how to account for the motion by the camera itself by explicitly feeding information about the ego motion, such as velocity and rotation rate, into the network. We illustrate the proposed task and final approach in Figure~\ref{fig:overview}.

\parsection{Contributions} Our main contributions are the following:\newline \textbf{(i)} We propose the  task of future object detection -- a simple, yet challenging, extension of traditional object detection. We establish multiple reference methods for this task on the publicly available nuImages and nuScenes~\cite{caesar2020nuscenes} datasets.\newline
\textbf{(ii)} We design a spatiotemporal detection transformer, with an extensive analysis of different attention mechanisms that would enable incorporation of information from additional frames. We demonstrate that our approach significantly improves upon both the single-frame and spatiotemporal reference methods. \newline
\textbf{(iii)} We show that providing raw ego-motion information to the network leads to a large performance increase. We find that this provides an even larger benefit than extending the model to multiple frames, and that these effects are complementary.

%% file: inputs/related_work.tex
\section{Related Work}
\parsection{Prediction} Prediction of future video frames in terms of RGB-images has been previously studied with temporal convolution~\cite{kalchbrenner2017video}, adversarial learning~\cite{mathieu2016deep,liang2017dual}, recurrent neural networks~\cite{ranzato2014video,srivastava2015unsupervised,villegas2017decomposing}, parallel multi-dimensional units~\cite{byeon2018contextvp}, variational recurrent neural networks~\cite{castrejon2019improved}, atoms (from system identification)~\cite{liu2018dyan}, and spatial transformers~\cite{liu2017video}. This problem is not only interesting in its own right, but enables unsupervised learning of video representations~\cite{liang2017dual}. The works of Villegas \etal~\cite{villegas2017learning} and Wu \etal~\cite{wu2020future} demonstrated that such video prediction may benefit from explicit modelling of dynamic objects. More closely related to our work is future prediction in terms of more high-level concepts, such as trajectories of dynamic objects. This task has been previously tackled using detect-track-predict paradigm~\cite{chai2020multipath,hong2019rules,tang2019multiple,styles2020multiple}. Hu \etal~\cite{hu2021fiery} demonstrated the benefits of instead learning this task end-to-end. This was achieved via a differentiable warping~\cite{jaderberg2015spatial} of feature maps to a reference frame with the help of ego motion. The warped feature maps were then processed with spatiotemporal 3D-convolutions. Perhaps most closely related to our work is the work of Luc \etal~\cite{luc2018predicting}, where future instance segmentations are predicted in an end-to-end fashion. To this end, MaskRCNN~\cite{he2017mask} was modified with a convolutional feature forecasting module. An advantage of future object detection, compared to future instance segmentation, is that the annotations are more commonly used and cheaper, typically resulting in larger datasets. Moreover, the dataset used by Luc \etal did not include ego-motion information. Li \etal~\cite{li2020towards} noted that the latency in object detectors provide a need of future prediction. The focus of their work was to analyze this aspect and they found that classical tracking-based approaches outperform the end-to-end trained neural network proposed by Luc \etal. The concurrent work by Peri \etal~\cite{peri2022forecasting} proposes 3D future object detection for LiDAR data. In contrast to their work, we instead investigate camera-based future object detection directly in the image plane.

\parsection{Object Detection with Transformers} The work of Carion \etal~\cite{carion2020end}, DETR, showed how object detection can be learnt end-to-end. They proposed to let a representation of the desired object detection output -- a set of object slots -- cross-attend a representation of the input -- a feature map extracted from the image. A concurrent work~\cite{chi2020relationnet++} showed how the cross-attention mechanism could be used to move between different output representations used in prior works, \eg, anchor-boxes of FasterRCNN~\cite{ren2015faster}, corners of CornerNet~\cite{law2018cornernet}, and center points of FCOS~\cite{tian2019fcos} or CenterNet~\cite{zhou2019objects}. Since then, several works have improved the effectiveness of transformer-based detectors. Dai \etal~\cite{dai2021up} proposed a pre-training stage. Wang \etal~\cite{wang2021pnp} proposed a multi-scale variant. Zhu \etal~\cite{zhu2020deformable} replaced the attention layers with multi-head, multi-feature-level deformable convolution. Dai \etal~\cite{dai2021dynamic} investigated another form of multi-feature-level deformable convolution. Gao \etal~\cite{gao2021fast} improved the locality of the cross-attention heads via a Gaussian-like weighting. Meng \etal~\cite{meng2021conditional} proposed to change how the transformer makes use of its positional encodings. Amongst the most important effects of these works is the reduction in training time, compared to DETR. The work of Meng \etal, for instance, reports a massive reduction: an entire order of magnitude. 

\parsection{Processing Spatiotemporal Data with Transformers} A number of works extended the use of transformers into the 3-dimensional video domain. Girdhar \etal~\cite{girdhar2019video} lets region-of-interest-pooled feature vectors cross-attend a video feature map, extracted by I3D~\cite{carreira2017quo}. The spatiotemporal structure, \ie \emph{where} in the image and at \emph{what} point in time, is captured in the positional encodings. Jaegle \etal~\cite{jaegle2021perceiver} proposed Perceiver, a general neural network architecture able to process data with different structure. Similar to the work of Girdhar \etal, the spatiotemporal structure is captured via positional encodings and a cross-attention mechanism lets an output representation gather information from the input data. In their work, Perceiver is used to process 2D feature maps extracted from images, 3D feature maps extracted from videos, and point clouds. Wang \etal~\cite{wang2021end} use a similar mechanism for video instance segmentation (VIS). Duke \etal~\cite{duke2021sstvos} experiments with spatiotemporal self-attention for video object segmentation (VOS), and propose to use local attention to reduce computational cost in long videos. In contrast to action recognition or point cloud classification, both VOS and VIS require output that is dense both spatially and in time. Meinhardt \etal~\cite{meinhardt2021trackformer} uses a DETR-style transformer for detection and tracking. This is achieved via a recurrent connection, where the output representation in one frame is reused in the next.

In this work, we extend DETR to spatiotemporal data. We experiment with several different mechanisms able to process spatiotemporal data, as illustrated in Figure~\ref{fig:encoders}: joint cross-attention \cite{girdhar2019video,jaegle2021perceiver,wang2021end}, recurrent cross-attention \cite{meinhardt2021trackformer}, and a sequential cross-attention mechanism where previous points in time are processed with different transformer layers. Furthermore, we investigate the addition of ego-motion information to the model. Instead of using ego-motion and geometry to warp feature maps as in the work of Hu \etal~\cite{hu2021fiery}, we let a series of cross-attention layers learn how to best make use of this information.

%% file: inputs/method.tex
\section{Method}\label{sec:method}
We aim to design and train a neural network to make predictions about the future set of objects in the image plane. First, we formalize the task of future object detection. Next, we introduce a single-frame baseline -- based on DETR~\cite{carion2020end} -- that can be trained directly for this task. Then, we extend this model to be able to process spatiotemporal data, \ie videos, and capture the dynamics of the scene. Last, we also add  ego-motion information to our proposed neural network. Such information is often readily available from on-board sensors and we hypothesize that this is a powerful cue for the neural network to utilize.

\subsection{Future Object Detection}
Consider a sequence of images $\{\mathbf{I}^t\}_{t=1}^T$, where the image $\mathbf{I}^t\in\mathbb{R}^{3\times H_0\times W_0}$ was captured time $t$. The aim for a future object detector, $F$, is to, given the sequence of images, find the set of objects that are visible in a future frame,
\begin{align}
  \mathcal{O}^{T+\tau} = F(\mathbf{I}^1,\dots,\mathbf{I}^T)\enspace.
\end{align}
There exists an image $\mathbf{I}^{T+\tau}$ in which the objects $\mathcal{O}^{T+\tau}$ are visible but the future object detector does not observe this image. Similar to traditional object detection, each object $O_m^{T+\tau}\in\mathcal{O}^{T+\tau}$ is described by a vector of class probabilities $\mathbf{c}_m^{T+\tau}\in\mathbb{R}^{C}$ and a bounding box $\mathbf{b}_m^{T+\tau}\in\mathbb{R}^4$. In contrast to traditional object detection, future object detection strives to predict, or \emph{detect}, objects in a future \emph{unseen} image. If the prediction horizon, $\tau$, is zero, future object detection degenerates into video object detection. It should be noted that for real-time systems where objects need to be detected, $\tau$ is typically non-zero due to the lag introduced by the camera and object detector~\footnote{See Li \etal~\cite{li2020towards} for a detailed analysis of this problem.}.

The future object detection formulation has two appealing properties. First, in order to predict $\mathcal{O}^{T+\tau}$, it is necessary to recognize and understand the dynamics present in $\{\mathbf{I}^t\}_{t=1}^T$. A traditional pipeline tackling this task would typically detect objects, track these objects, and extrapolate their trajectories. Conceptually, these are the steps we think that a future object detection method would need. The second step, object tracking, typically requires densely annotated sequences. However, with this formulation, only the future frame on which prediction is to be made, corresponding to time $T+\tau$, needs to be annotated. 

Second, the object detection formulation provides a principled way of encoding the uncertainty of the future. Given that the presence of an object in $\{\mathbf{I}^t\}_{t=1}^T$ has been determined, it is typically uncertain exactly where this object ends up at time $T+\tau$. The object detection formulation enables the placement of multiple bounding boxes, each corresponding to a different hypothesis for the trajectory of that object. The number of bounding boxes is variable, making the method free to decide on the number of plausible hypotheses. Without any modification, standard object detection losses~\cite{he2017mask,carion2020end} reward the model for making such multi-modal predictions. We therefore adopt the standard set criterion~\cite{carion2020end} when training models for the inherently multi-modal future object detection task.

\subsection{Detection Transformer Baseline}
Let us first consider the adaptation of a standard object detector to future object detection. This can be achieved by training the detector, $F$, to predict
\begin{align}
    \mathcal{O}^{T+\tau} = F(\mathbf{I}^T)\enspace.
    \label{eq:baseline}
\end{align}
Again, note that traditional object detection is retrieved in the limit $\tau\rightarrow 0$. For other values $\tau > 0$, the input $\mathbf{I}^T$ may not always permit $\mathcal{O}^{T+\tau}$ to be retrieved. For instance, the velocities of objects are unknown and consequently also their location at time $T+\tau$. However, we expect the detector $F$ to make good guesses. Furthermore, even when accurate predictions are impossible, we expect the detector to accurately estimate the uncertainty. Common detectors~\cite{he2017mask,carion2020end} are designed and trained to achieve this by returning multiple hypotheses (bounding boxes) with reduced confidence.

In theory, any method for object detection could be trained to solve \eqref{eq:baseline}. We focus on the transformer-based family of object detection methods, first proposed by Carion \etal~\cite{carion2020end}. These methods are composed of a CNN backbone, a transformer encoder, and a transformer decoder. First, we encode the image $\mathbf{I}^T$ with the backbone, providing a deep feature map $\mathbf{x}^T\in\mathbb{R}^{D\times H\times W}$ of lower resolution, $H<H_0, W<W_0$. The deep feature map is then processed by a transformer encoder~\cite{vaswani2017attention},
\begin{align}
    \mathbf{z}^T = \enc(\mathbf{x}^T)\enspace.
    \label{eq:enc}
\end{align}
The transformer encoder comprises a set of self-attention layers. Each feature vector $\mathbf{x}_{i,j}^T\in\mathbb{R}^D$ attends all feature vectors in the feature map $\mathbf{x}^T$. This provides the capacity to add global context and model long-range relationships, which provides a rich representation of the image, $\mathbf{z}^T\in\mathbb{R}^{D\times H\times W}$. Next, a transformer decoder, $\dec$ produces a set of $M$ object representations. The decoder contains learnable queries $\mathbf{y}^\text{init}\in\mathbb{R}^{M\times D}$. These queries gather information from the encoded feature maps,
\begin{align}
    \mathbf{y}^T = \dec(\mathbf{y}^\text{init}, \mathbf{z}^T)\enspace.
\end{align}
Each query $\mathbf{y}_m^T\in\mathbb{R}^D$ corresponds to one possible object, with the prediction based on the input at time $T$. Finally, a head predicts the box and class scores for each query,
\begin{align}
    \mathbf{c}_m^{T+\tau} = \classhead(\mathbf{y}_m^T)\enspace, \quad   \mathbf{b}_m^{T+\tau} = \boxhead(\mathbf{y}_m^T)\enspace.
\end{align}
Each prediction $(\mathbf{c}_m^{T+\tau}, \mathbf{b}_m^{T+\tau})$ corresponds to one hypothesis at a future frame $T+\tau$. We train the approach to predict the set of objects at time $T+\tau$ using the matching based loss introduced in DETR~\cite{carion2020end} (details in Appendix~\ref{appendix:implementation-details}).

\subsection{Capturing Dynamics with Spatiotemporal Mechanisms}\label{sec:spatiotemporal}
Making predictions about the future requires information about the dynamics. A classical pipeline would typically detect objects in multiple frames and then create tracks from the detections, in order to estimate their dynamics. Based on the estimated dynamics, predictions can be made about the future. In some scenarios, it may be possible to guess the dynamics from a single image -- for instance, vehicles  usually follow their lane with a speed equal to the speed limit -- but in the general case, a future object predictor requires information about the dynamics. We therefore extend DETR to process sequences of images.

We investigate three straightforward extensions to the attention mechanism in DETR: joint attention, sequential cross-attention, and recurrent transformer. Each of these three mechanisms enables DETR to process sequences of images. The joint attention mechanism lets the encoder or decoder jointly attend all image feature maps, $[\mathbf{x}^1,\dots,\mathbf{x}^T]$ or $[\mathbf{z}^1,\dots,\mathbf{z}^T]$ respectively. the sequential cross-attention mechanism attends each image feature map with a different attention layer. Last, we experimented with a recurrent transformer. The recurrent transformer interleaves attending the image feature maps and the output at the previous time step, either $\mathbf{z}^{T-1}$ or $\mathbf{y}^{T-1}$ depending on whether the encoder or decoder is made recurrent. In our experiments, we found sequential cross-attention to yield the best performance, and we describe it next. For details on the joint attention and recurrent transformer, see Appendix~\ref{appendix:spatiotemporal-mechanism}.

\parsection{Sequential Cross-Attention} We replace the single self-attention operation or the single cross-attention operation in $\enc$ or $\dec$, respectively, with a sequence of attention operations. For the transformer encoder, consider one of the $L$ transformer encoder layers, $\enc^i$. This layer contains a self-attention operation,
\begin{align}
    \mathbf{z}^{T,i,0} = \enc^{i,0}(\mathbf{z}^{T,i-1}, \mathbf{z}^{T,i-1})\enspace.
\end{align}
We additionally cross-attend each previous feature map $\{\mathbf{x}^t\}_{t=1}^{T-1}$,
\begin{align}
    \mathbf{z}^{T,i,k} = \enc^{i,k}(\mathbf{z}^{T,i,k-1}, \mathbf{x}^{T-k}),\quad k=1,\dots,T-1\enspace.
    \label{eq:sca-enc-2}
\end{align}
Here, $\enc^{i,k}$ is a single attention layer with its own set of weights. We assign $\mathbf{z}^{T,0}=\mathbf{x}^T$ and $\mathbf{z}^{T,i} = \mathbf{z}^{T,i,T-1}$. The final output is $\mathbf{z}^T=\mathbf{z}^{T,L}$. Note that \eqref{eq:sca-enc-2} relies on $\{\mathbf{x}^t\}_{t=1}^{T-1}$, requiring the backbone to be run on each input image $\{\mathbf{I}^t\}_{t=1}^T$.

Sequential cross-attention in the decoder follows a similar set of equations. A single decoder layer $\dec^i$ comprises two steps. First, a self-attention layer is applied to the queries,
\begin{align}
    \mathbf{y}^{T,i,0} = \dec^{i,0}(\mathbf{y}^{T,i-1}, \mathbf{y}^{T,i-1})\enspace.
\end{align}
The queries then sequentially cross attend the input encodings,
\begin{align}
    \mathbf{y}^{T,i,k} = \dec^{i,k}(\mathbf{y}^{T,i,k-1}, \mathbf{z}^{T-k}),\quad k=1,\dots,T\enspace.
    \label{eq:sca-dec-2}
\end{align}
We assign $\mathbf{y}^{T,0}=\mathbf{y}^\text{init}$ and $\mathbf{y}^{T,i} = \mathbf{y}^{T,i,T}$. The final output is obtained as $\mathbf{y}^{T}=\mathbf{y}^{T,L}$. In \eqref{eq:sca-dec-2}, we rely on the availability of $\{\mathbf{z}^t\}_{t=1}^{T-1}$, and both the backbone and encoder needs to be applied to all images $\{\mathbf{I}^t\}_{t=1}^T$.

Both these operations exhibit linear complexity in the sequence length, $T$. Input from each point in time is processed with its own set of weights, enabling the neural network to treat feature vectors captured at different points in time differently. This added flexibility enables the neural network to search for different patterns at different time steps and to use retrieved information differently. We do not encode the temporal position of each feature vector. To the best of our knowledge, this mechanism has not been previously studied, despite its simplicity. It is somewhat reminiscent of axial attention~\cite{ho2019axial,wang2020axialdeeplab} where each axis, \eg, height, width, and time, is processed separately.

\subsection{Incorporating Ego-motion Information}\label{sec:egomotion}
A major challenge in future object detection is that the motion of dynamic objects is highly correlated with the \emph{ego-motion} of the camera. Autonomous robots often move and even objects that are stationary in the three-dimensional sense could move drastically in the image plane. It is possible to rely on multiple images to estimate and compensate for the ego-motion using visual odometry (VO) and geometry. In principle, the neural network could learn VO and use it as an internal representation. However, many systems are equipped with additional sensors that capture ego-motion information. We hypothesize that such information could constitute a powerful cue for the network to use.

While the exact sensor setups and ways of extracting ego motion can vary between different robots, we believe it is reasonable to assume that one has access to at least some subset of position, velocity, acceleration, rotation and rotation rate. Such information may be three-dimensional, but in the case of autonomous vehicles it could be projected to the road plane. For maximum flexibility, and generalization to different setups, we make minimal assumptions on which ego-motion information is available. Our only processing is to convert information into a local coordinate system. For example, global position or rotation is converted into a transformation relative to the last frame. We concatenate all available information and pass it through an ego-motion encoder,
\begin{equation}
    \mathbf{f}_{\text{ego}} = \phi([\text{translation}, \text{speed}, \text{rotation}, ...])\enspace,
\end{equation}
where $[\cdot,\cdot]$ denotes vector concatenation. We adopt a 2-layer MLP as our ego-motion encoder, $\phi$.

Following the philosophy in Section~\ref{sec:spatiotemporal}, we wish to use cross-attention to incorporate additional sources of information, such as the encoded ego-motion vector. However, it should be noted that cross-attending a single vector degenerates to a simple linear projection and addition -- significantly reducing the computational cost. Nevertheless, letting $\mathbf{z}^{t,i}$ denote the output of $\enc^i$ corresponding to time $t$, we can set

\begin{align}
    \mathbf{z}^{t,i} = \enc^{i,\text{ego}}(\mathbf{f}_\text{ego}^t, \mathbf{z}^{t,i})\enspace.
\end{align}
Correspondingly, in the decoder, ego-motion can be incorporated by setting
\begin{align}
    \textbf{y}^{t,i} = \dec^{i,\text{ego}}(\mathbf{f}_\text{ego}^t, \mathbf{y}^{t,i})\enspace.
\end{align}
Both of these operations enables the neural network to process ego-motion information in a completely learning based manner.

%% file: inputs/experiments.tex
\section{Experiments}
We implement Future Object Detection training, evaluation, multiple reference methods, and the proposed spatiotemporal mechanisms. We use PyTorch~\cite{paszke2019pytorch} and experiment on the NuScenes and NuImages datasets~\cite{caesar2020nuscenes}. See Appendix~\ref{appendix:implementation-details} for additional details. The code will be made publicly available to facilitate future research on future object detection or for other tasks with the models used in this work. First, we report quantitative results for the proposed spatiotemporal mechanisms and compare them to multiple references. Second, we conduct two ablation studies. Third, we provide qualitative results. Additional experimental results are available in Appendix~\ref{appendix:additional-experimental}.

\subsection{Dataset}
Although we need video datasets, there is no need for the entire video to be annotated. Both training and evaluation is possible with a single frame of the video being annotated. To this end, we experiment with NuScenes~\cite{caesar2020nuscenes} and the subsequent dataset NuImages. NuImages contains samples of 13 images collected at 2 Hz; ego-motion information for each image; and object detection annotations for the 7th image. To simplify the problem we only consider front camera images, leaving us with 13066 training samples and 3215 validation samples. NuScenes is primarily intended for 3D object detection using the full sensor suite, and provides 3D annotations at 2Hz. However, projected 2D bounding boxes are also available. NuScenes is inconsistent in the camera sampling frequency, since the 20Hz camera is resampled to 12Hz. During evaluation, we discard samples which do not match the desired sampling frequency exactly. NuImages and NuScenes has two major advantages for future object detection -- advantages that are natural for datasets aimed towards autonomous robots. First, each dataset is captured with a single sensor suite, avoiding for instance issues with different cameras. Second, ego-motion information is included, which we hypothesize is a powerful cue for future object detection. .

\subsection{Quantitative Results}

\begin{table*}[t]
    \centering
    \resizebox{0.8\linewidth}{!}{%
    \begin{tabular}{l c c c c c c c c c}
        \toprule
        \multirow{2}{*}{\textbf{Method}} & \multicolumn{3}{c}{\textbf{AP50}} & \multicolumn{3}{c}{\textbf{AP50}} & \multicolumn{3}{c}{\textbf{AP}}\\
         & Mean & Car & Pedestrian & Small & Medium & Large & Mean & Car & Pedestrian\\
        \midrule
        Oracle                       & 67.5 & 87.2 & 73.5 & 42.0 & 69.5 & 84.7 & 38.4 & 56.2 & 35.7\\
        \midrule
        Naïve                        &  9.8 & 14.5 &  2.5 &  3.6 &  9.4 & 18.8 &  3.0 &  4.2 &  0.6\\
        Tracking                     & 14.8 & 16.9 &  3.6 &  3.3 & 14.8 & 25.9 &  4.5 &  5.0 &  0.8\\
        \midrule
        FasterRCNN~\cite{ren2015faster}        & 15.6 & 24.4 &  6.1 &  2.5 & 15.2 & 31.3 &  4.8 &  7.5 &  1.5 \\
        ConditionalDETR~\cite{meng2021conditional}          & 13.5 & 22.5 &  6.2 &  3.0 & 12.9 & 30.8 & 3.8 &  6.7 &  1.5\\
        F2F adaptation~\cite{luc2018predicting} & 21.6 & 35.6 & 10.8 &  3.3 & 22.0 & 42.0 &  7.2 & 12.2 & 2.7\\
        \midrule
        Spatiotemporal (Ours)        & 23.9 & 36.6 & 12.0 & 4.1 & 25.1 & 52.1 & 8.1 & 12.1 &  3.1 \\
        Singleframe + Ego (Ours)     & \second{25.6} & \second{38.1} & \second{12.8} & \first{5.1} & \second{26.9} & \second{54.8} & \second{8.9} & \second{13.2} & \second{3.4} \\
        Spatiotemporal + Ego (Ours)  & \first{28.0} & \first{43.2} & \first{14.7} & \second{4.7} & \first{29.9} & \first{59.3} & \first{9.4} & \first{15.5} &  \first{3.8}\\
        \bottomrule
    \end{tabular}
    }
    \caption{Performance for future object detection on the nuImages validation set in terms of AP50 and AP (higher is better). The prediction horizon is 500 ms. The best performing entry is marked in \first{red} and second best in \second{blue}.}
    \label{tab:baseline-comparison}
    \vspace{-8mm}
\end{table*}

We begin by exploring the task of future object detection using a naïve baseline, a tracking-based method, an oracle, and several adaptions of prior art. The naïve baseline corresponds to the scenario where no future prediction is done. We run standard object detection on the past image ($\mathbf{I}^T$) and treat the detections as the future prediction. The tracking-based method follows the baseline used in FIERY~\cite{hu2021fiery}: The centerpoint distance between pairs of detections in two different frames is used as cost and then the Hungarian algorithm is used to minimize the total cost. Matched objects are extrapolated linearly into the future frame. The oracle acts as a best-case reference, providing the performance of a single-frame detector applied directly to the future frame ($\mathbf{I}^{T+\tau}$). We hypothesise that it could be beneficial to learn future object detection directly, in and end-to-end manner. To that end, we adapt two existing object detectors: FasterRCNN~\cite{ren2015faster}, and ConditionalDETR. In both cases, the detector is presented with the past image ($\mathbf{I}^T$) and trained to predict the future objects ($\mathcal{O}^{T+\tau}$). We also compare with F2F~\cite{luc2018predicting}, which the authors show to outperform multiple baselines -- including optical flow warping -- on the closely related task of future instance prediction. To adapt it to future object detection, and simplify comparisons with our method, we use a DETR-like approach but replace the transformer encoder with the F2F module, which takes features from multiple frames and learns to forecast a future feature map. These reference methods are compared to our proposed method, which extends DETR with: a spatiotemporal mechanism, using $T=2$; ego-motion information; or both. In our initial experiments, we found $T>2$ to provide negligible gain at high computational cost. We leave making efficient use of additional frames to future work.

The results are shown in Table~\ref{tab:baseline-comparison}. The naive approach fares very poorly, indicating large changes from $T$ to $T+\tau$. The tracking solution improves the performance across the board, but still struggles. Likely, the simple matching and extrapolation heuristics prove insufficient over the large prediction horizon. Both FasterRCNN and ConditionalDETR perform comparably with the tracking-based method despite only having access to a single frame, showing the potential of end-to-end training for this task. The spatiotemporal F2F adaptation proves much stronger, in several cases more than twice as good as the tracking solution. Our proposed spatiotemporal method consistently outperforms F2F, especially for large objects, likely due to the flexibility of our attention mechanism. Interestingly, both spatiotemporal methods are outperformed by our Singleframe + Ego solution - clearly demonstrating the advantage of incorporating ego-motion. Our final approach, incorporating both multiple frames and ego-motion outperforms the other approaches for all but the smallest objects. However, we note that there is still a substantial gap to the oracle, likely due to the stochastic nature of the future.

\begin{table}[t]
    \centering
    \resizebox{0.7\linewidth}{!}{%
    \begin{tabular}{l c c c c c c c c}
        \toprule
        \multirow{2}{*}{\textbf{Method}} & \multicolumn{2}{c}{\textbf{50 ms}} & \multicolumn{2}{c}{\textbf{100 ms}} & \multicolumn{2}{c}{\textbf{250 ms}} & \multicolumn{2}{c}{\textbf{500 ms}}\\
        & Car & Ped. & Car & Ped. & Car & Ped. & Car & Ped.\\
        \midrule
        Oracle                      & 71.4 & 49.9 & 69.7 & 46.5 & 72.1 & 49.9 & 72.3 & 49.8\\
        \midrule
        Naïve                       & 70.4 & 41.7 & 64.6 & 25.6 & 47.0 & 12.1 & 27.4 & 4.9\\
        Tracking                    & 70.9 & 43.6 & 66.1 & 36.6 & 50.6 & 23.6 & 30.9 & 10.0\\
        Singleframe + Ego (Ours)    & 71.3 & 47.7 & 69.6 & 44.1 & 62.9 & 33.3 & 50.9 & 17.8\\
        Spatiotemporal + Ego (Ours) & \textbf{72.4} & \textbf{48.3} & \textbf{70.1} & \textbf{44.2} & \textbf{65.6} & \textbf{35.8} & \textbf{54.9} & \textbf{21.5}\\
        \bottomrule
    \end{tabular}
    }
    \caption{Performance for future object detection on the NuScenes validation set in terms of AP50 (higher is better) for multiple prediction horizons.}
    \label{tab:baseline-latencies}
    \vspace{-6mm}
\end{table}

Part of the motivation behind future prediction is to compensate for the latency induced by the object detector. For the relatively long horizon of 500 ms, compared to typical network latencies, we have seen that the zero-latency oracle has much better performance. Table \ref{tab:baseline-latencies} compares the performance of our future predictions with the oracle, for a number of prediction horizons. As expected, the gap shrinks for shorter horizons. More surprisingly, we find a threshold, near 100 ms, where the future predictions are roughly on par with the oracle. In the case of cars, we even outperform the oracle, likely thanks to the addition of ego-motion and spatiotemporal cues. These results suggests an additional element to the traditional latency-accuracy tradeoff -- the future prediction horizon.

\begin{table}[t]
    \centering
    \resizebox{0.7\linewidth}{!}{%
    \begin{tabular}{l c c c c}
        \toprule
        \multirow{2}{*}{\textbf{Method}} & \multicolumn{2}{c}{\textbf{AP50}} & \multicolumn{2}{c}{\textbf{AP}} \\
         & Car & Pedestrian & Car & Pedestrian \\
        \midrule
        Singleframe                 & 22.5          & 6.2           & 6.7           & 1.5          \\
        Joint Attention Encoder     & 34.8          & 10.5          & 11.8          & 2.7          \\
        Joint Attention Decoder     & \second{35.4} & \second{11.0} & \second{11.9} & \second{2.8} \\
        Sequential CA Encoder       & 35.0          & 9.7           & \second{11.9} & 2.5          \\
        Sequential CA Decoder       & \first{36.6}  & \first{12.0}  & \first{12.1}  & \first{3.1}  \\
        Recurrent Tr. Encoder       & 34.9          & 9.5           & 11.8          & 2.4          \\
        Recurrent Tr. Decoder       & 32.6          & 9.2           & 10.8          & 2.3          \\
        \bottomrule
    \end{tabular}
    }
    \caption{Future object detection performance on the NuImages validation set for different spatiotemporal mechanisms.}
    \label{tab:ablations}
    \vspace{0mm}
´\end{table}

\parsection{Spatiotemporal Ablation} In Section~\ref{sec:method}, we introduced three mechanisms that each enables the neural network to process video data. Each mechanism can be applied either in the encoder or the decoder. We compare the different alternatives and report the results in Table~\ref{tab:ablations}. Compared to the single-frame version, all spatiotemporal variants improve performance substantially. This indicates that each of the three different mechanisms learns to capture and make use of the dynamics in the scene. Interestingly, joint and sequential attention work best in the decoder, while the recurrent mechanism obtain the best results in the encoder. Best performance is obtained with sequential cross-attention in the decoder.

\begin{table}[t]
    \centering
    \resizebox{0.7\linewidth}{!}{%
    \begin{tabular}{l c c c c}
        \toprule
        \multirow{2}{*}{\textbf{Method}} & \multicolumn{2}{c}{\textbf{AP50}} & \multicolumn{2}{c}{\textbf{AP}} \\
         & Car & Pedestrian & Car & Pedestrian \\
        \midrule
        No ego-motion         & 36.6          & 12.0          & 12.1          & 3.1          \\
        Ego-motion in encoder & \first{43.2}  & \first{15.1}  & \first{15.5}  & \second{3.4} \\
        Ego-motion in decoder & \second{42.4} & \second{13.7} & \second{14.9} & \first{3.6}  \\
        \bottomrule
    \end{tabular}
    }
    \caption{Future object detection performance on the NuImages validation set with and without ego-motion.}
    \label{tab:egomotion}
\end{table}

\parsection{Ego-motion Ablation} Next, we introduce ego-motion information to our spatiotemporal neural network. In Section-\ref{sec:egomotion}, we described how to encode the ego-motion information and incorporate it into the network in a learnable manner, either in the encoder or the decoder. We evaluate the effectiveness of these approaches in Table~\ref{tab:egomotion}. As already established, ego-motion is a powerful cue that leads to significantly improved performance. However, the specific location where ego-motion is incorporated seems of minor importance. Here we choose to add it in the encoder. 

\begin{figure}[t]
  \centering
  \setlength{\tabcolsep}{2pt}
  \begin{tabular}{ccc}
    \includegraphics[width=.31\linewidth, trim=0mm 0mm 0mm 32mm, clip]{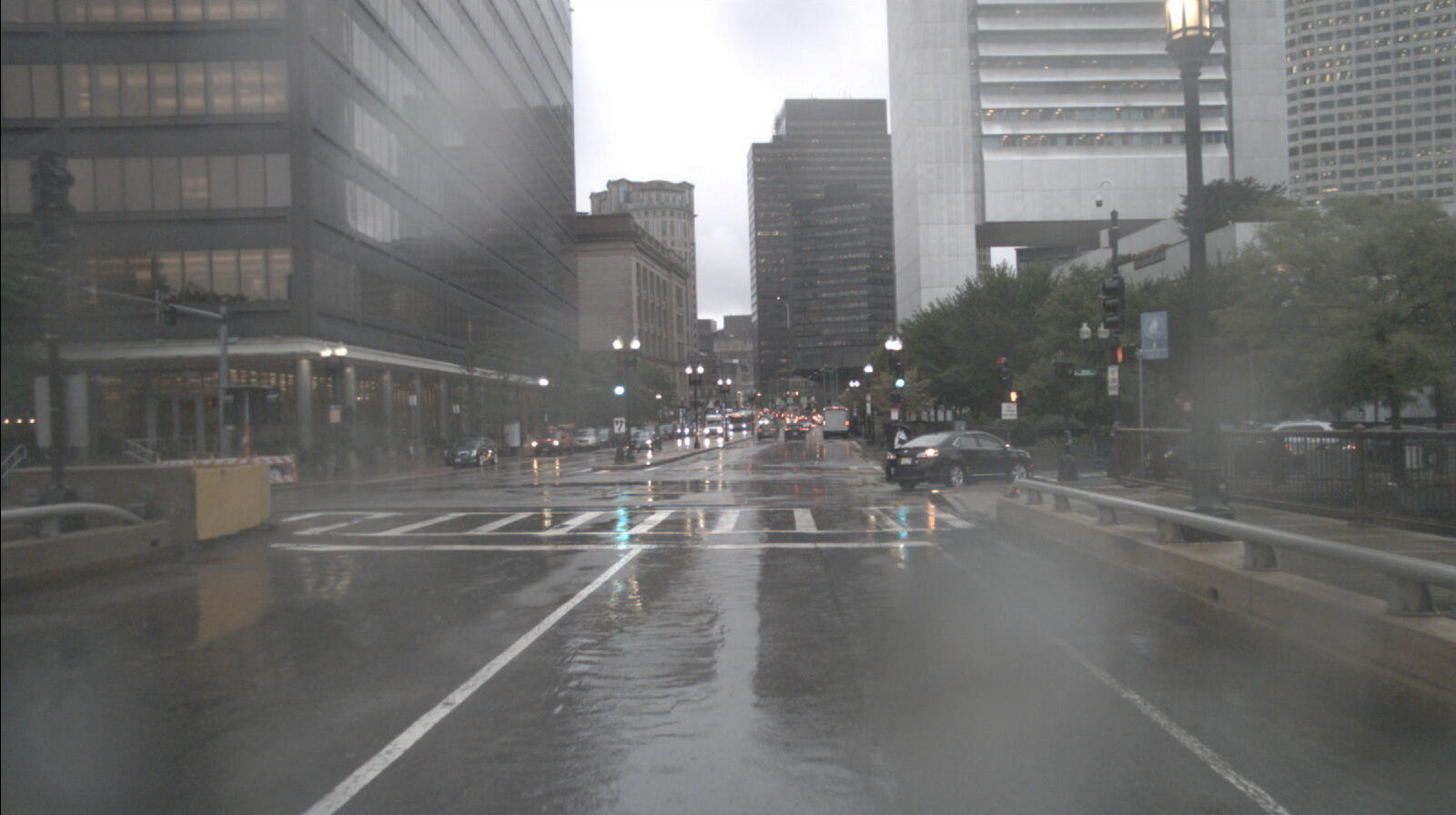} & \includegraphics[width=.31\linewidth, trim=0mm 0mm 0mm 32mm, clip]{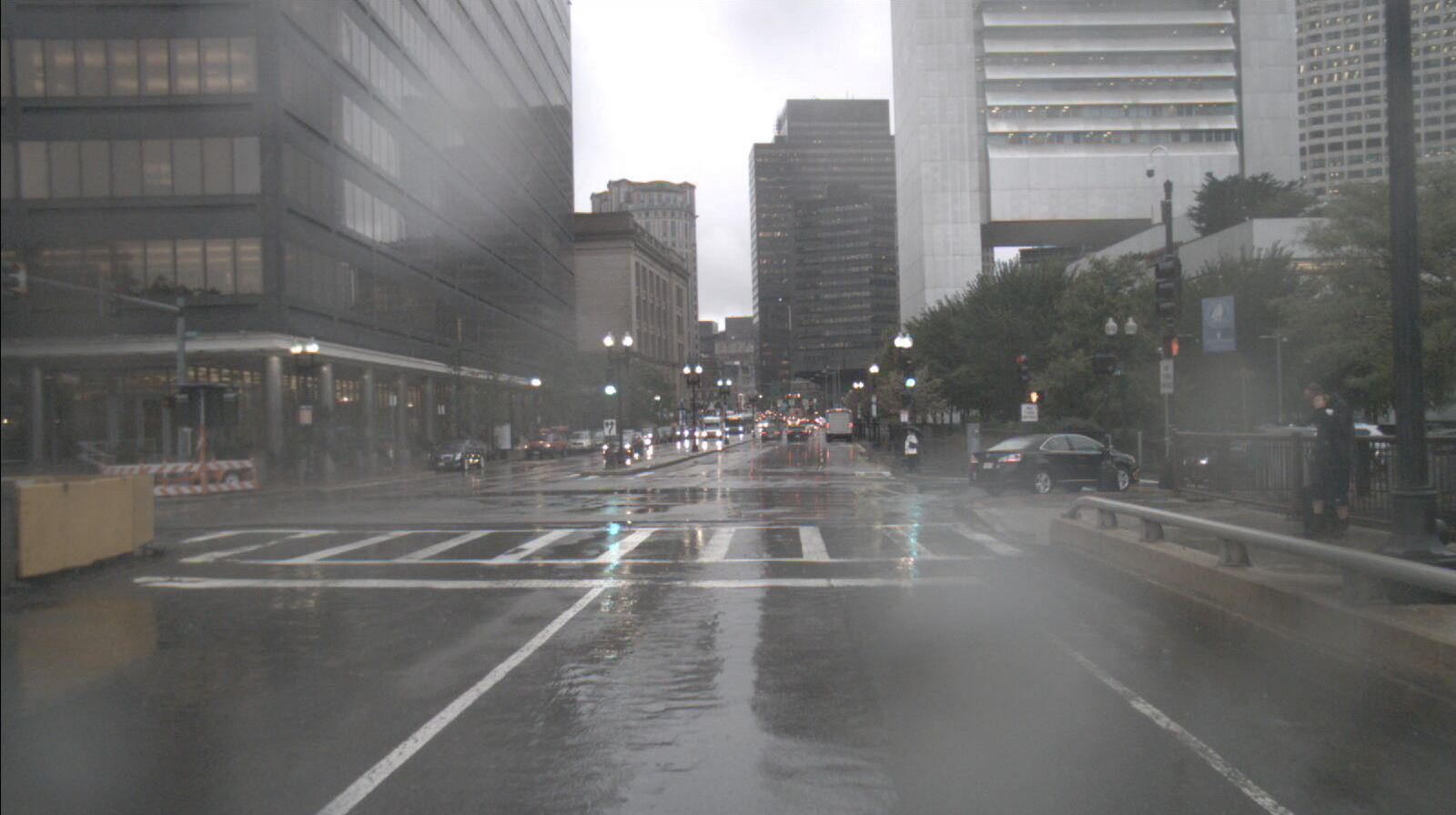} & \includegraphics[width=.31\linewidth, trim=0mm 0mm 0mm 32mm, clip]{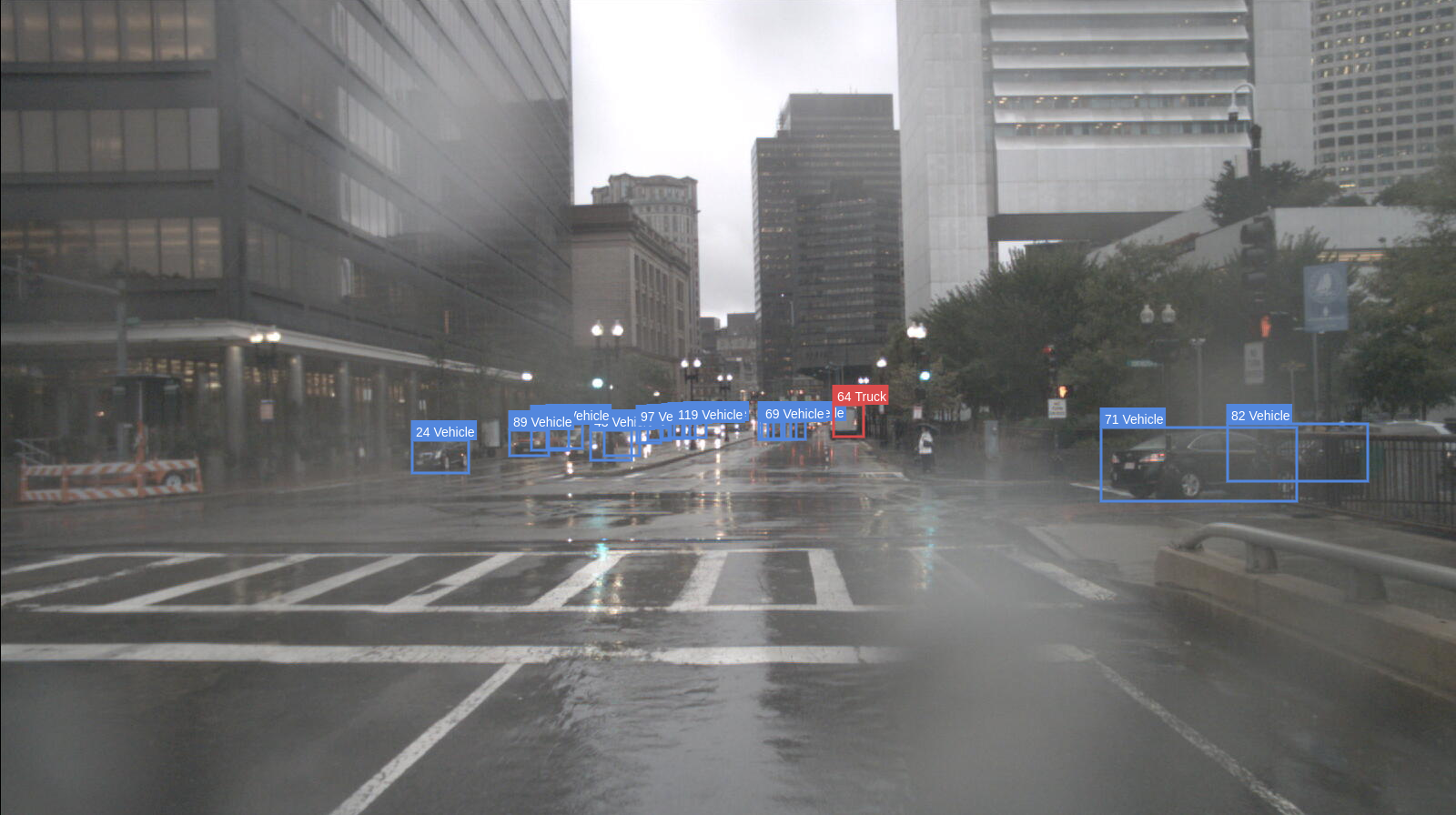}\\
    
    \includegraphics[width=.31\linewidth, trim=0mm 0mm 0mm 32mm, clip]{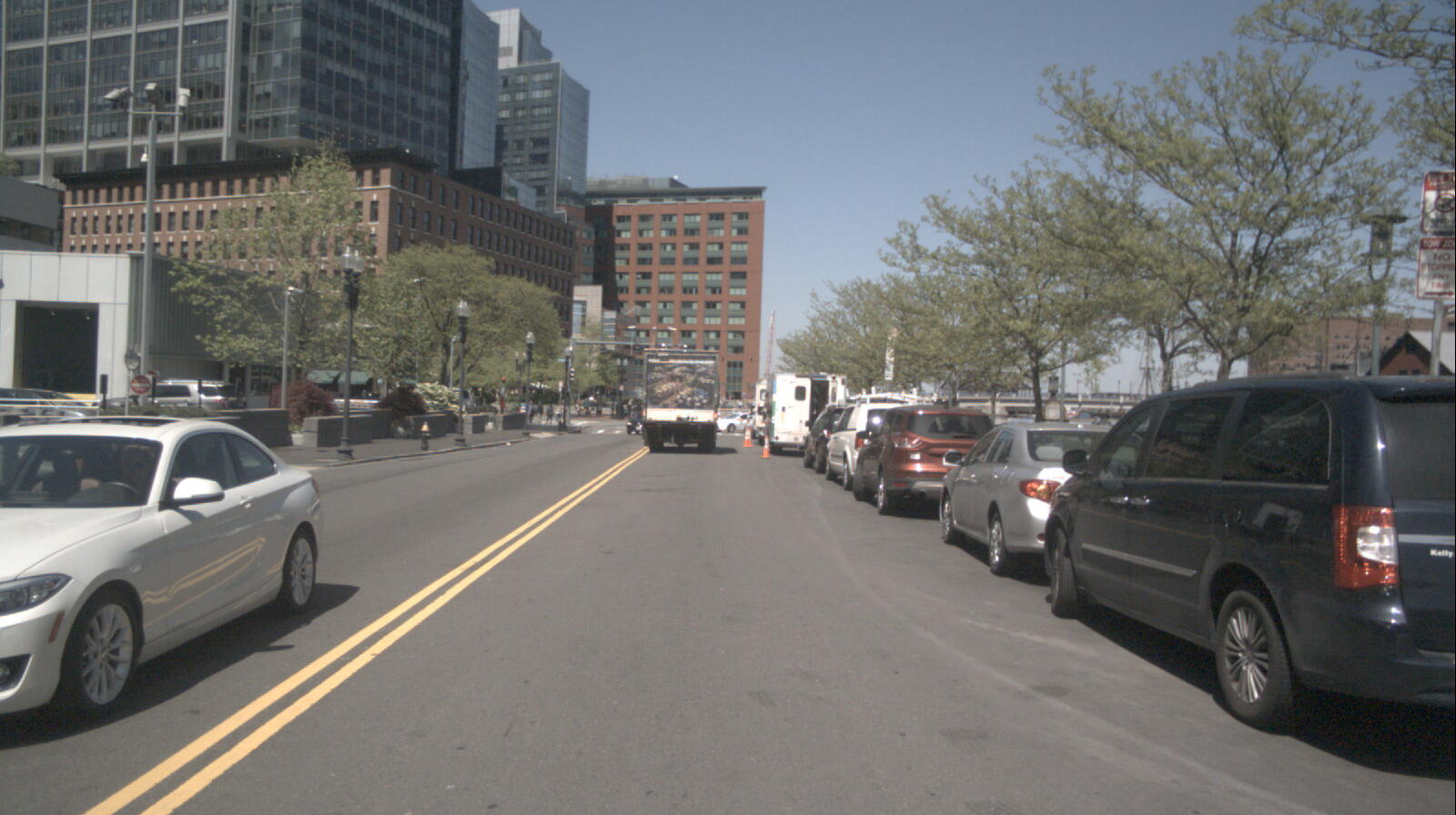} & \includegraphics[width=.31\linewidth, trim=0mm 0mm 0mm 32mm, clip]{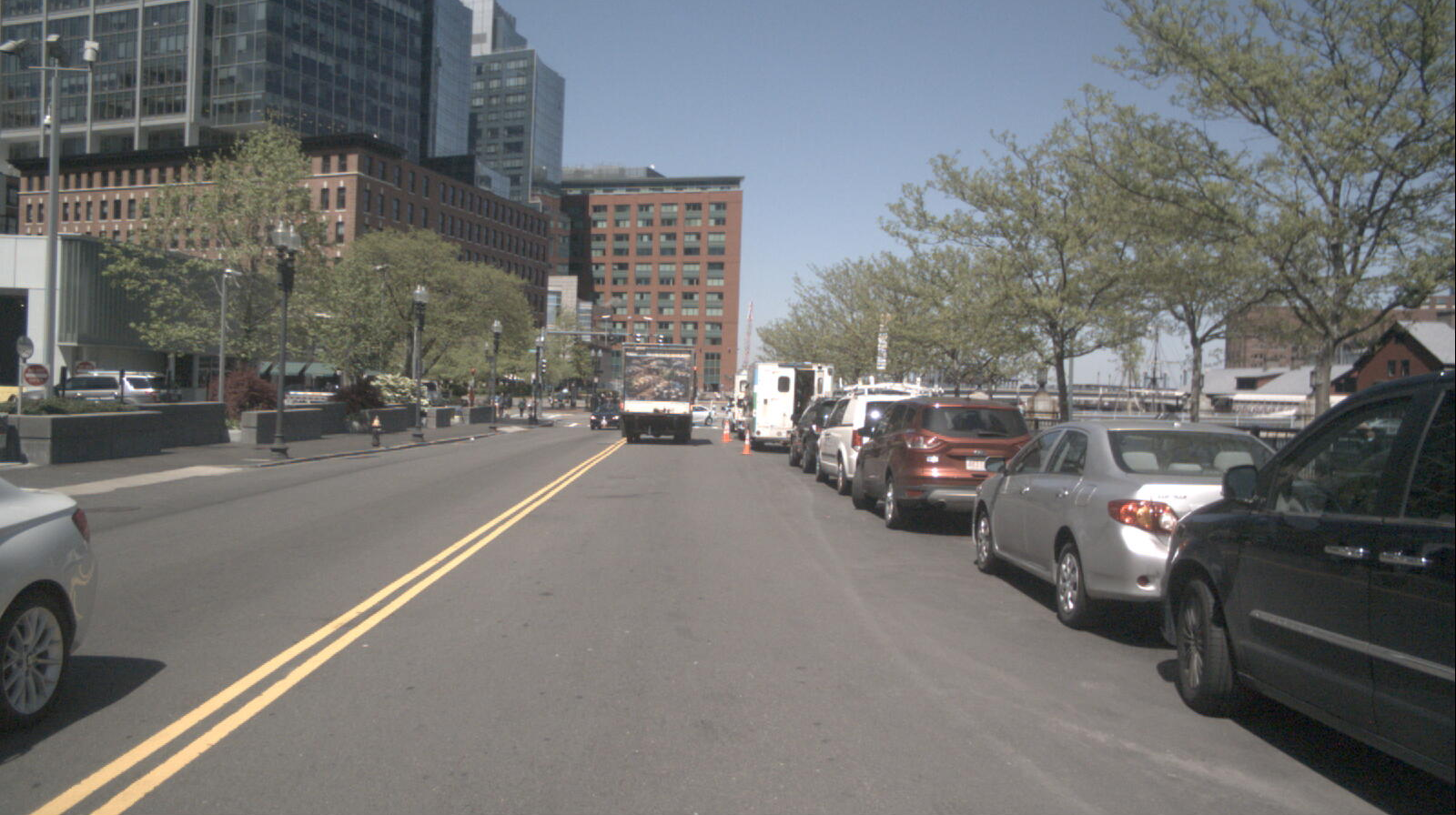} & \includegraphics[width=.31\linewidth, trim=0mm 0mm 0mm 32mm, clip]{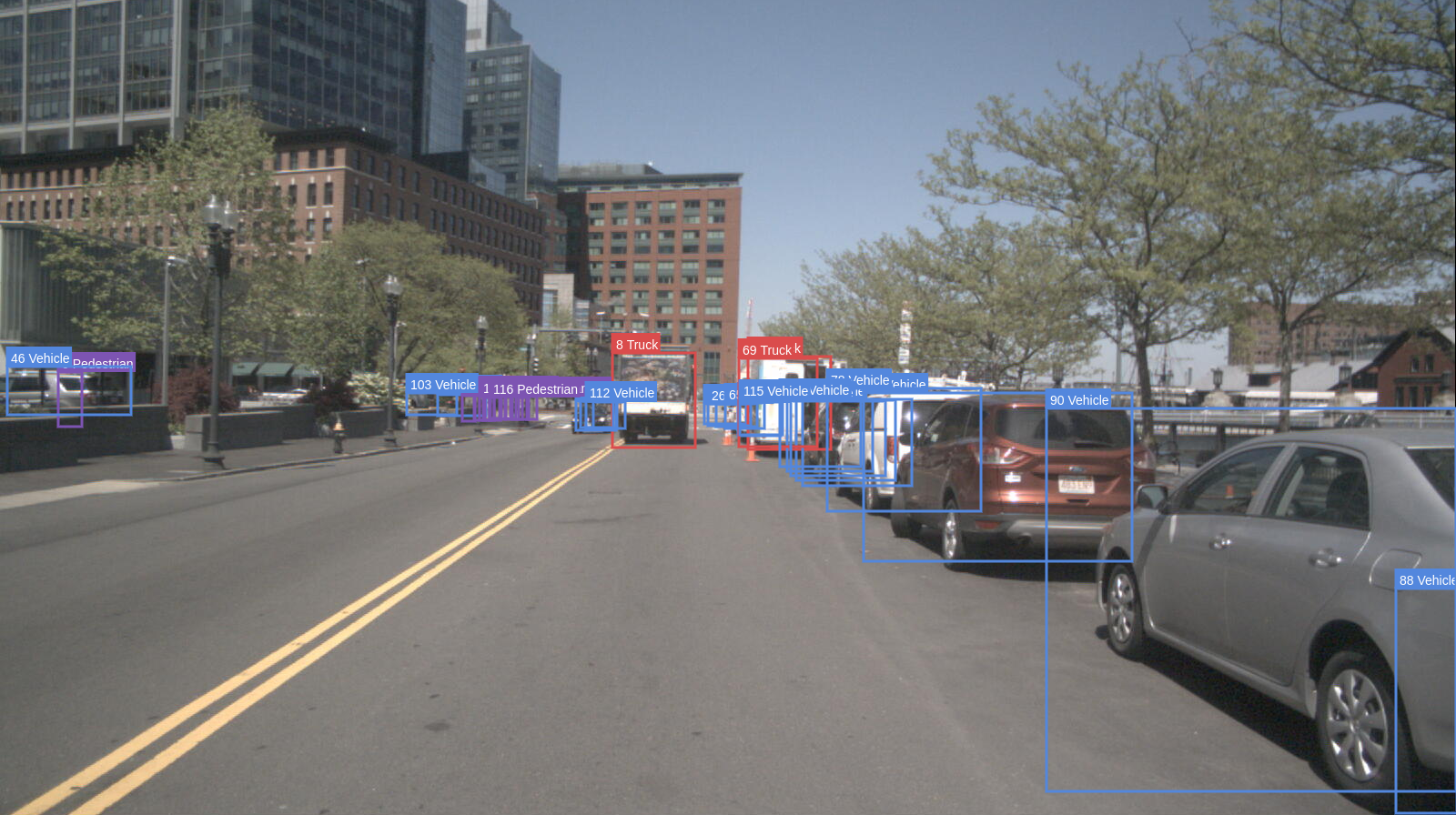}\\
    
    \includegraphics[width=.31\linewidth, trim=0mm 0mm 0mm 32mm, clip]{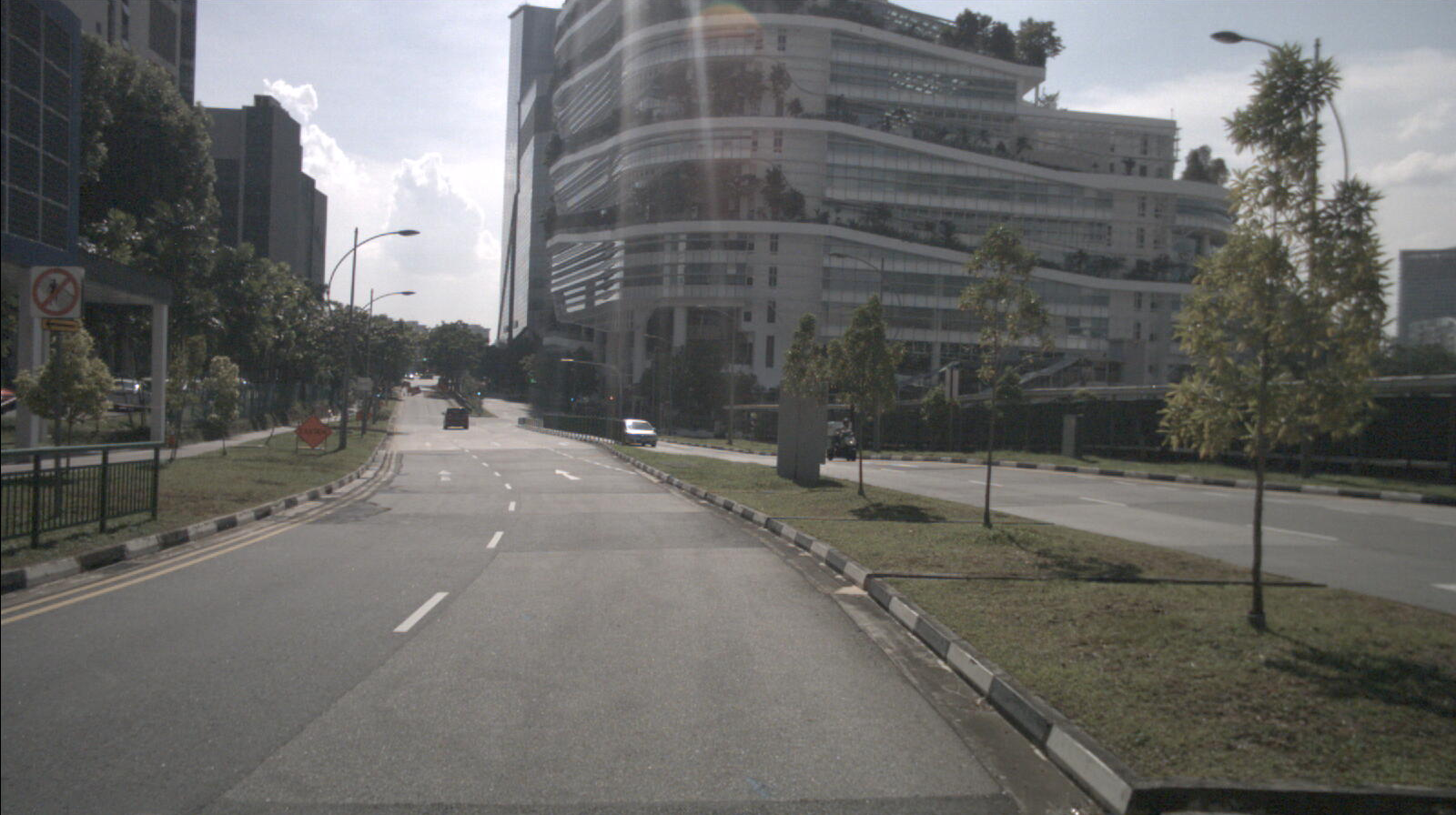} & \includegraphics[width=.31\linewidth, trim=0mm 0mm 0mm 32mm, clip]{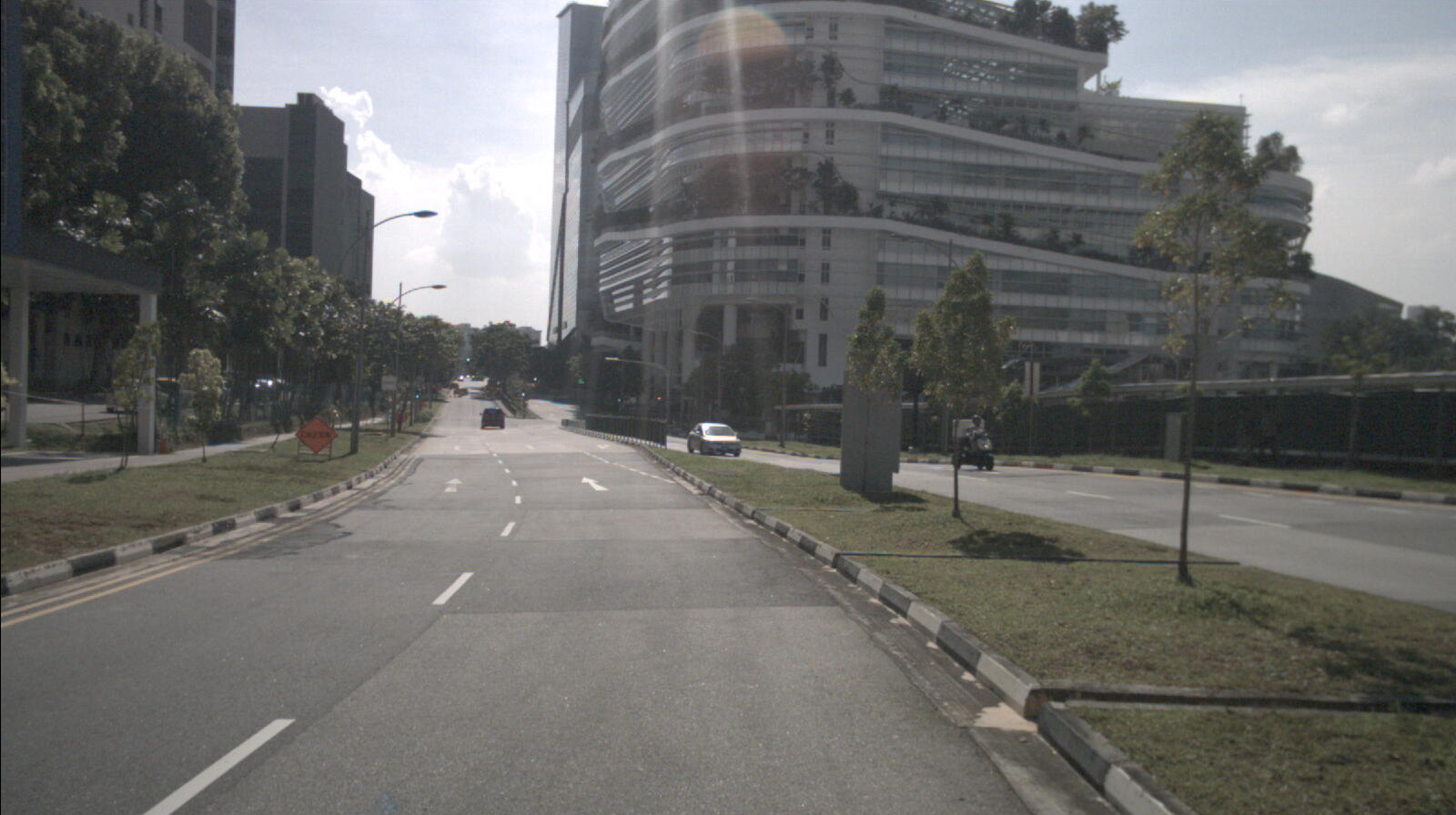} &
    \includegraphics[width=.31\linewidth, trim=0mm 0mm 0mm 32mm, clip]{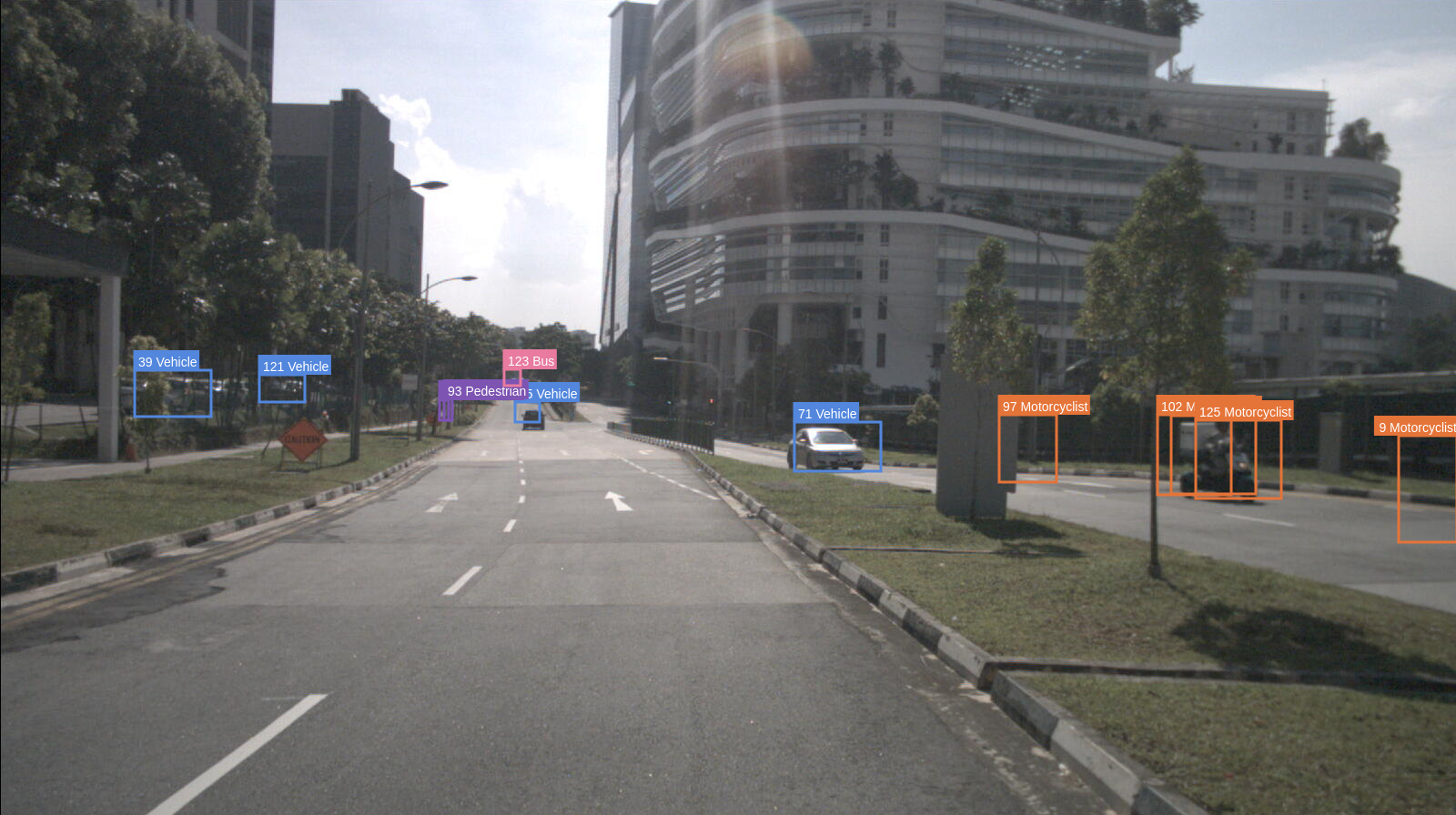}\\
  \end{tabular}
  \vspace{0mm}
  \caption{Three challenging sequences where our approach predicts the future. The left and center images are two consecutive frames, and the right-most image is the third, \emph{future} frame. Based on the two left-most frames, our approach predicts the objects in the third, future frame. In the third row, there are five motorcycle predictions. Most likely, the method is uncertain about the movement of the single motorcycle in the scene.}
  \label{fig:qualitative-standard}
  \vspace{0mm}
\end{figure}

\begin{figure*}[t]
     \centering
     \setlength{\tabcolsep}{2pt}
     \begin{tabular}{c c c}
     \includegraphics[width=0.32\textwidth, trim=0mm 0mm 0mm 16mm, clip]{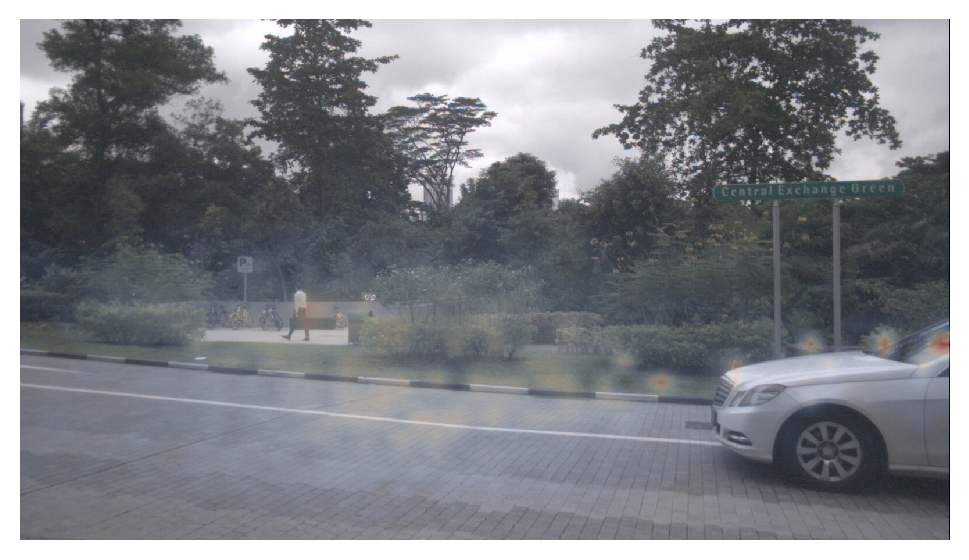} & \includegraphics[width=0.32\textwidth, trim=0mm 0mm 0mm 16mm, clip]{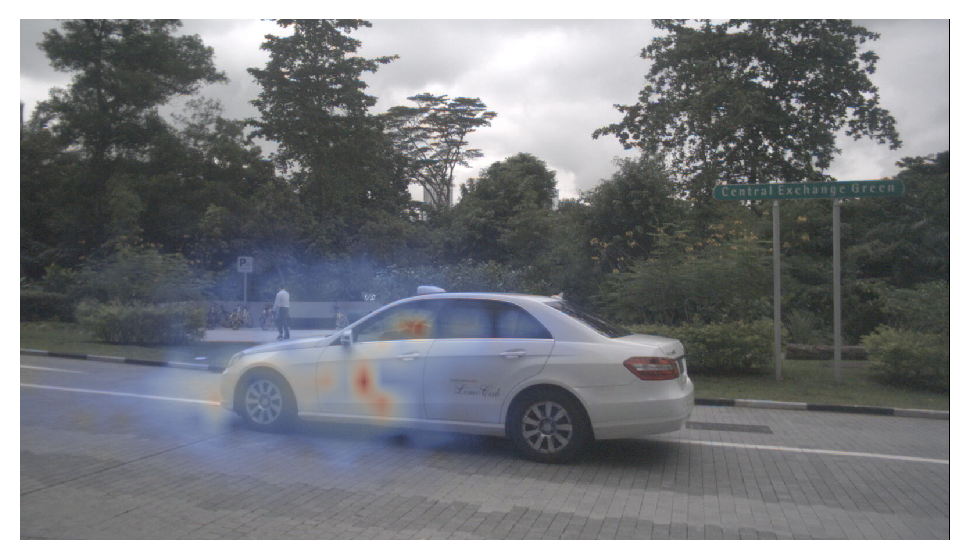} & \includegraphics[width=0.32\textwidth, trim=0mm 0mm 0mm 16mm, clip]{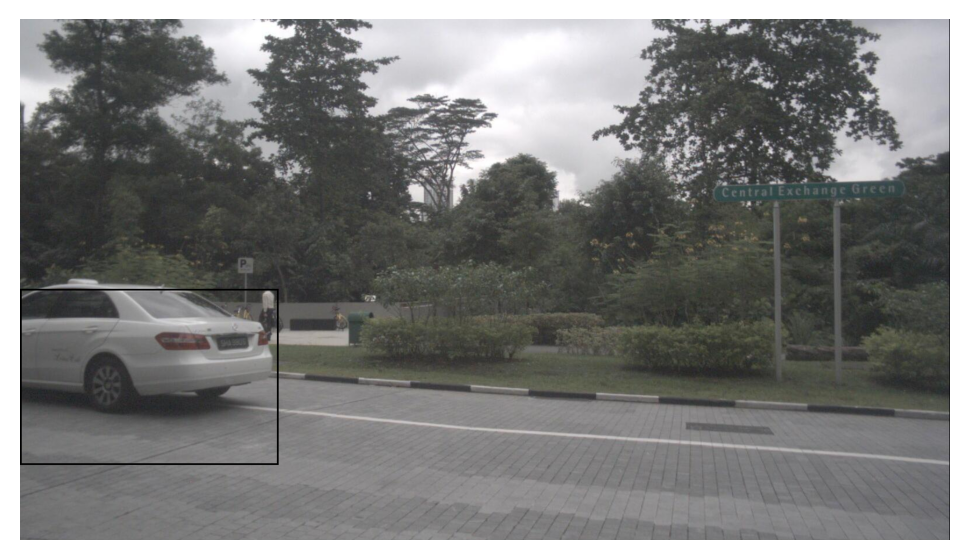}\\
     \includegraphics[width=0.32\textwidth, trim=40 45 115 80, clip]{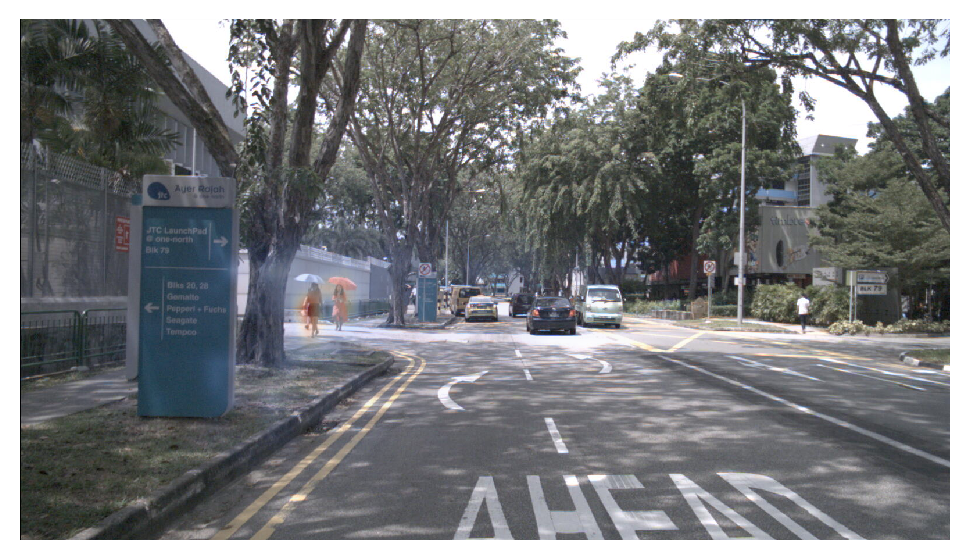} & \includegraphics[width=0.32\textwidth, trim=40 45 115 80, clip]{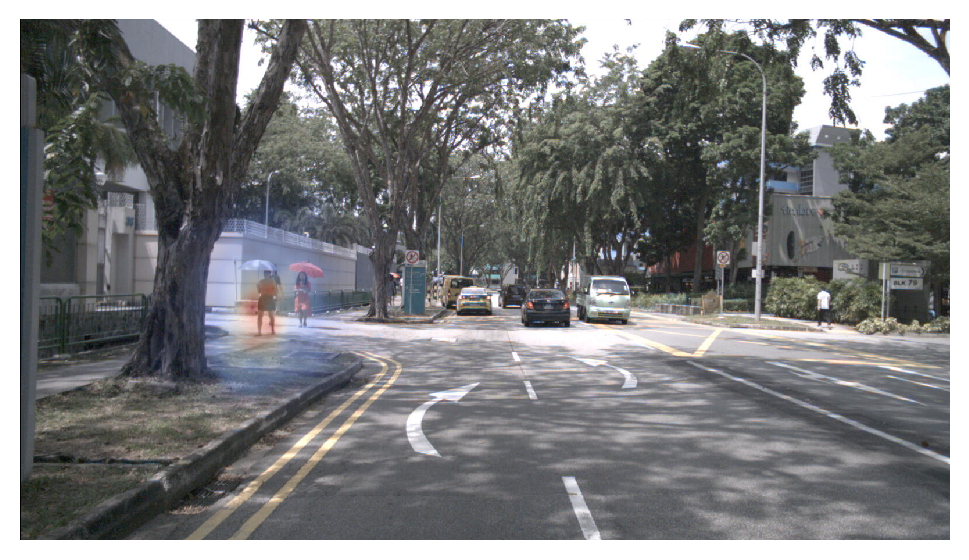} & \includegraphics[width=0.32\textwidth, trim=40 45 115 80, clip]{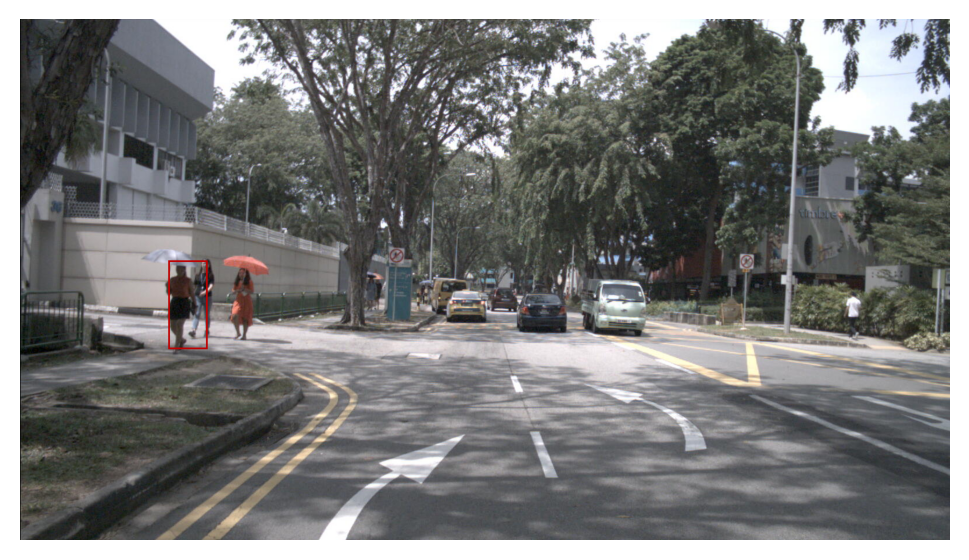}\\
     \end{tabular}
    \vspace{0mm}
    \caption{Visualization of a two future predictions with corresponding attention maps. The top row shows an object that moves swiftly across the image. The bottom row (zoomed) shows a low-visibility pedestrian that is standing still.}
    \label{fig:appendix_attention}
    \vspace{0mm}
\end{figure*}

\begin{figure*}[t]
  \centering
  \setlength{\tabcolsep}{2pt}
  \begin{tabular}{ccc}
    \includegraphics[width=.32\linewidth]{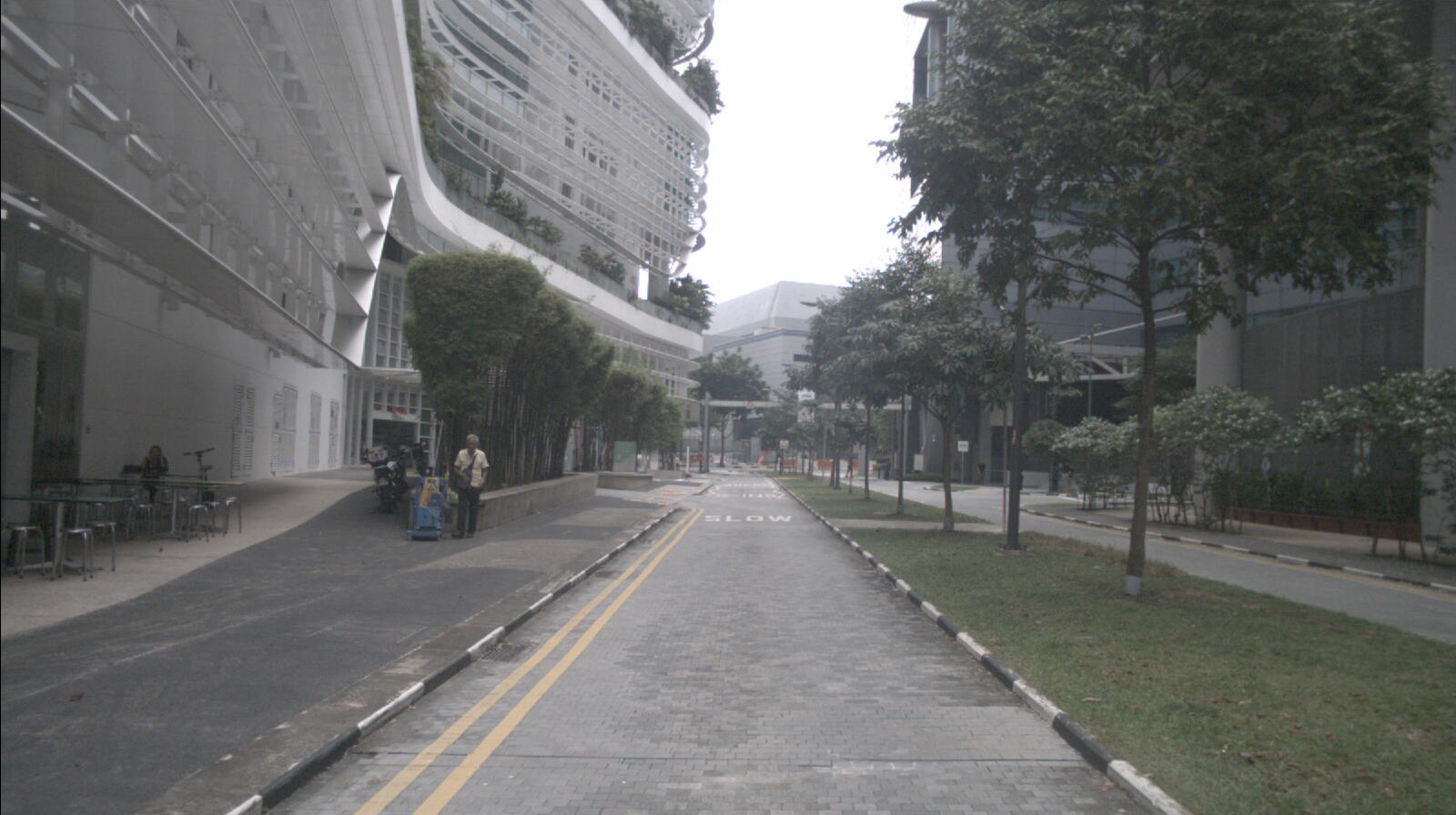} & \includegraphics[width=.32\linewidth]{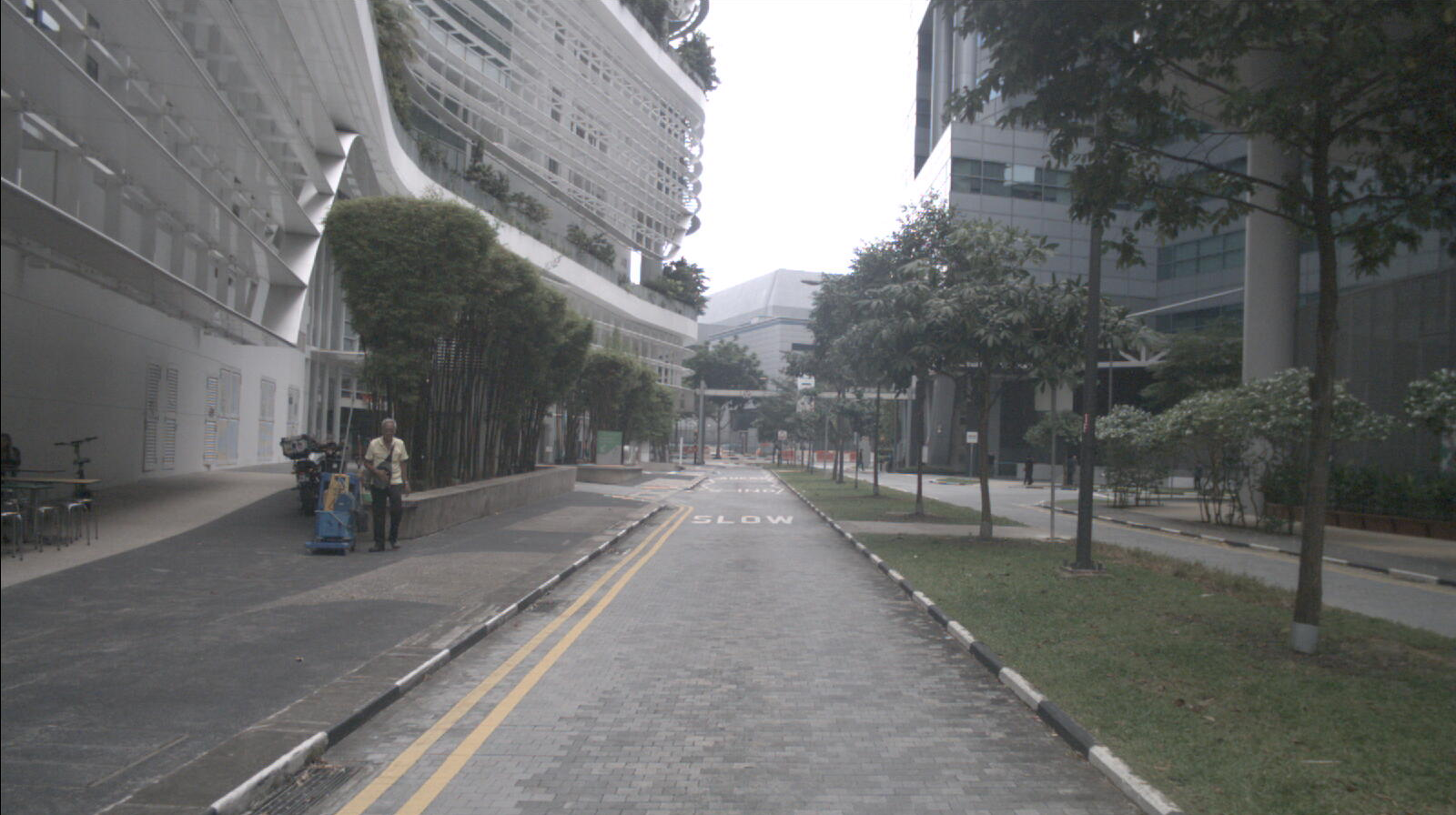} &
    \includegraphics[width=.32\linewidth]{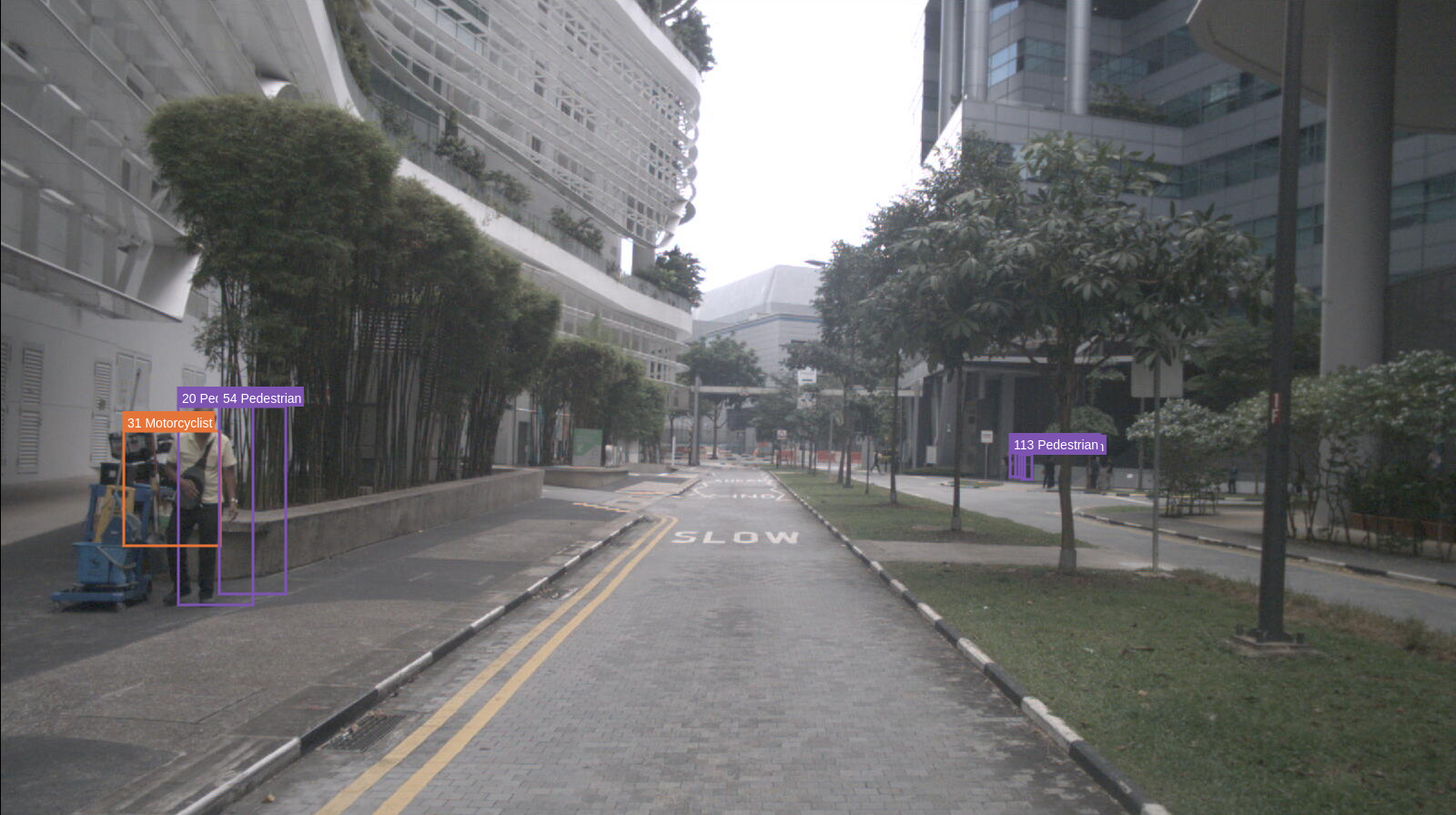}\\
    \includegraphics[width=.32\linewidth]{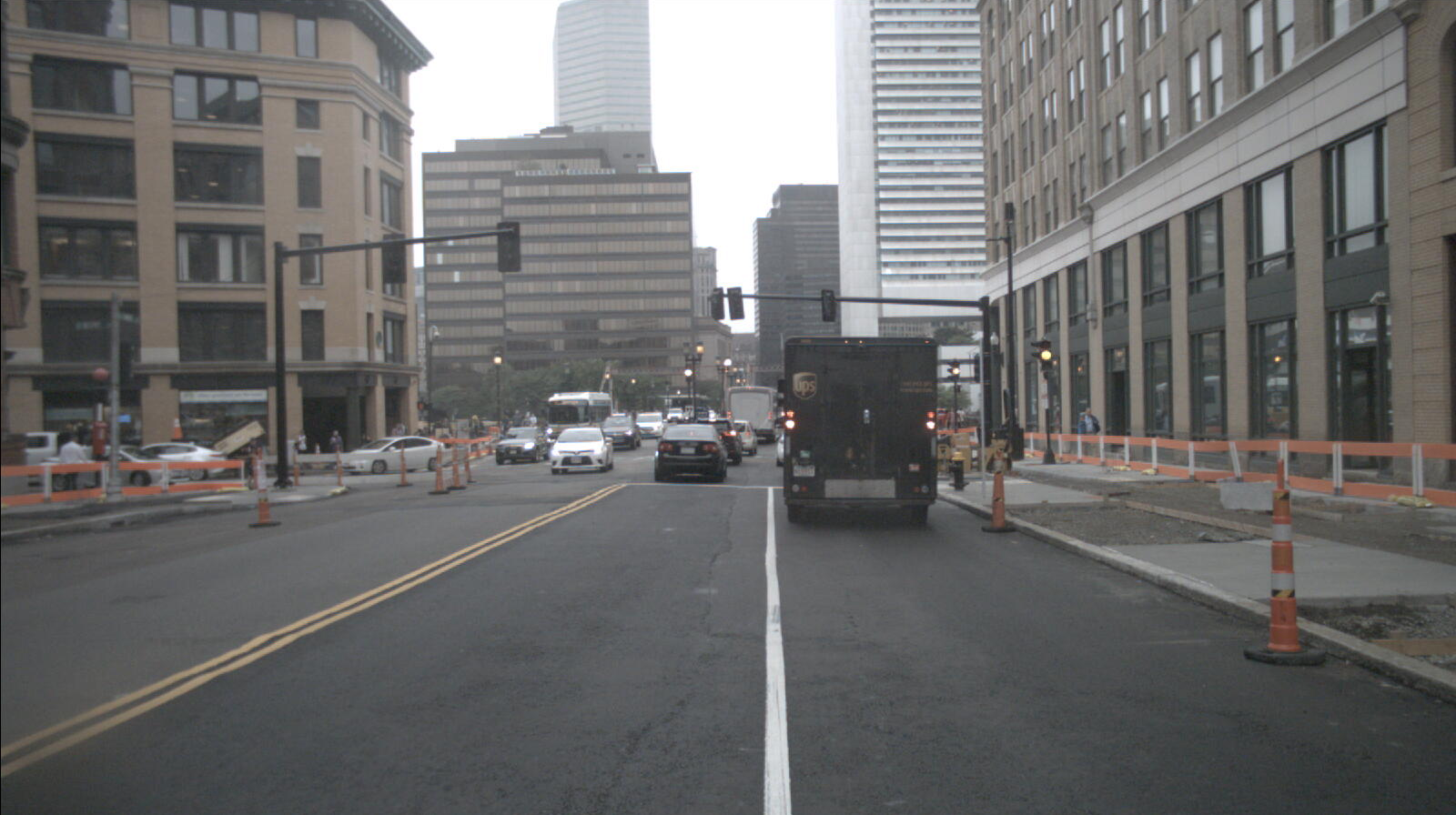} & \includegraphics[width=.32\linewidth]{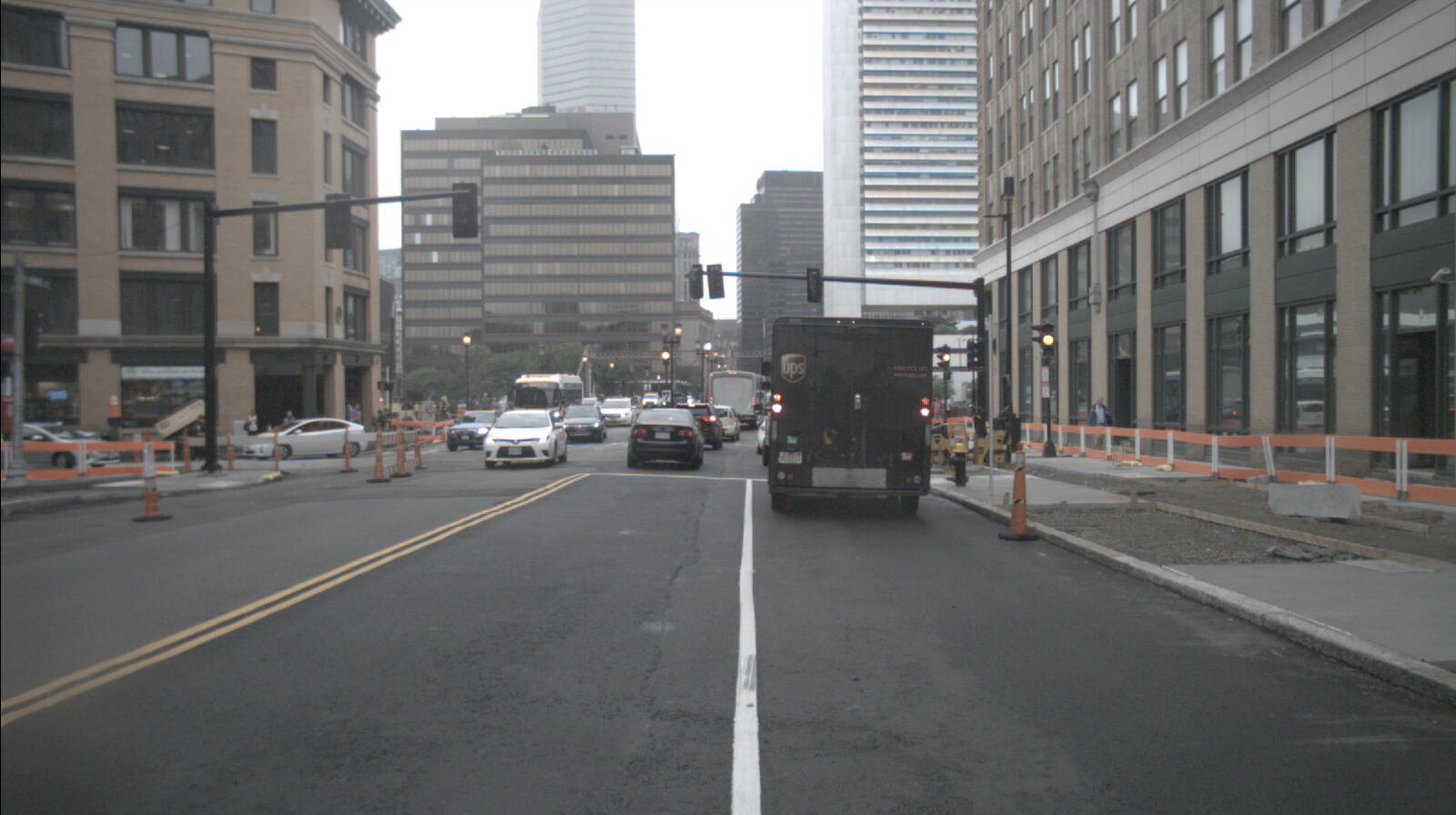} &
    \includegraphics[width=.32\linewidth]{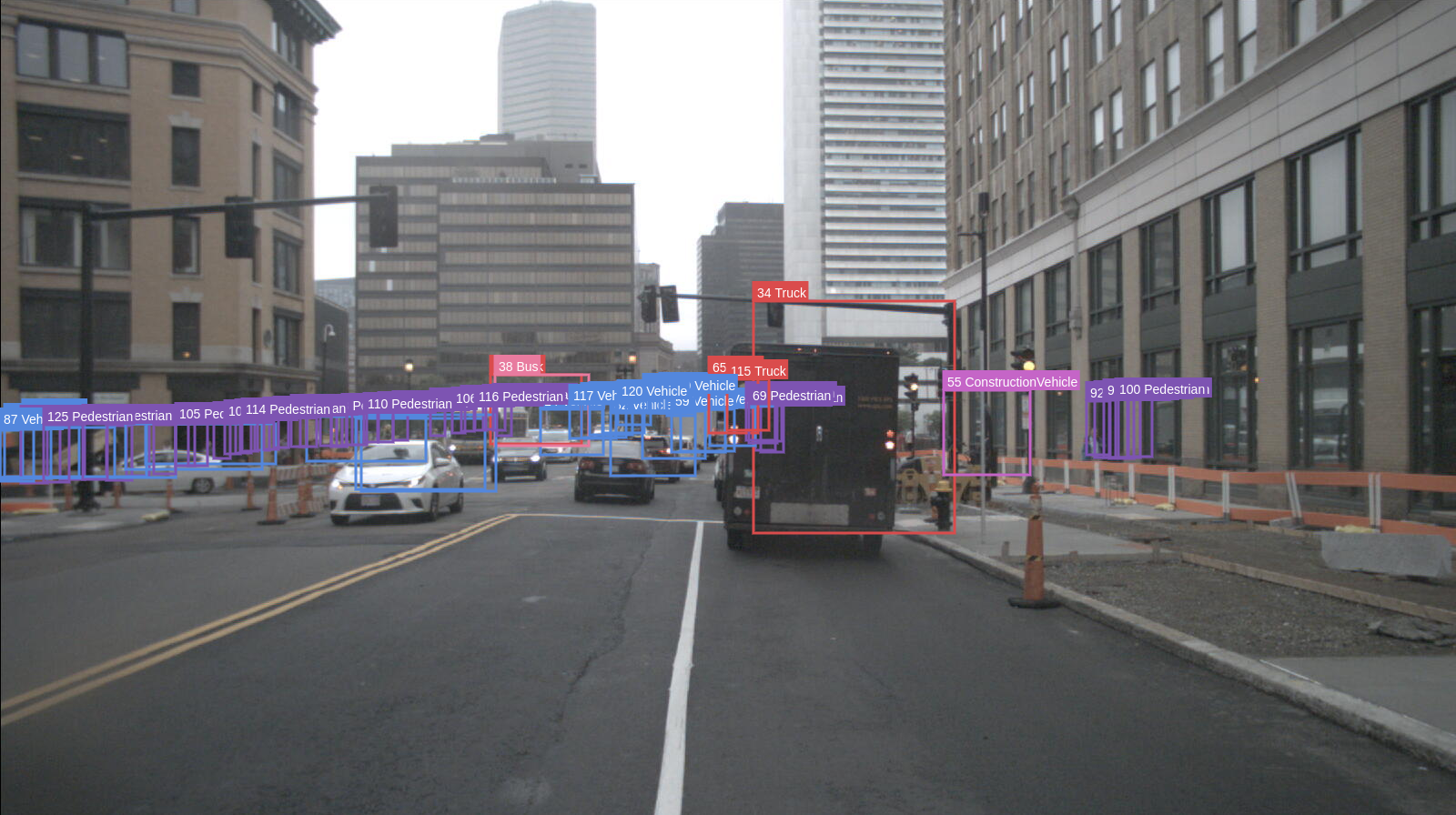}\\
  
  \end{tabular}
  \caption{Two example failure modes for our approach. On top, two objects are predicted for the pedestrian on the left side of the road. A possible explanation is that the model is uncertain about the exact location in the future, and predicts two hypotheses. On bottom, all predictions are off. In the video, there is a bump in the future frame. This is difficult for our model to capture.}
  \label{fig:qualitative-hard}
\end{figure*}

\subsection{Qualitative Results}
In Figure~\ref{fig:qualitative-standard}, we show qualitative results of our spatiotemporal approach on three challenging sequences. These three sequences exhibit ego-motion, moving objects, crowded scenes, and adverse weather condition. Even in such challenging conditions, the approach manages to accurately predict future object states. Figure~\ref{fig:qualitative-hard} shows two failure modes of our approach. In some scenarios, it may be impossible to accurately predict the future. In the top row, the neural network opts to predict multiple possible objects for a single human. This behaviour is common in object detectors in adverse conditions where they are uncertain. On the bottom row, all predictions have an offset as the car hits a bump in the road, an event that is difficult to foresee.

In Figure~\ref{fig:appendix_attention}, we visualize the attention maps for two additional sequences (also see Figure~\ref{fig:cool-future-pred}). In each sequence, we show a single predicted, future object. For the predicted object, we visualized the attention maps from the object slot to different regions in the two input images. The first sequence is especially difficult, since the object moves across the entire image. We can see that the model finds and focuses on the object in frame $T$, but spreads the attention across the entire image in frame $T-1$. The second sequence shows a number of pedestrians on the sidewalk. Even though they are relatively close to each other, the model finds and focuses on the correct individual in the past frames. 

%% file: inputs/conclusion.tex
\section{Conclusion}
We have proposed the task of future object detection, a simple extension of traditional object detection that challenges a model's ability to jointly predict the dynamics of moving objects and the camera itself. Through a variety of baselines and reference methods, we explored the capabilities of existing works on this task. We then developed an end-to-end spatiotemporal detection transformer that uses sequential cross-attention to process previous images as well as the raw ego-motion information. These improvements yielded major performance increases compared to the reference methods. Finally, we demonstrated qualitatively that a form of video tracking emerges in our model, even though it has only been trained with single-frame object annotations. We believe that this shows the potential of the proposed task formulation, and we hope that this work serves as a basis for future work on future object detection.

%% file: inputs/acknowledgements.tex
\parsection{Acknowledgements} This work was partially supported by the strategic research project ELLIIT and the the Wallenberg Artificial Intelligence, Autonomous Systems and Software Program (WASP) funded by Knut and Alice Wallenberg Foundation. The computations were enabled by resources provided by the Swedish National Infrastructure for Computing (SNIC), partially funded by the Swedish Research Council through grant agreement no. 2018-05973.

%% file: inputs/supplementary.tex
\clearpage
\appendix

\section{Additional Details}\label{appendix:implementation-details}
We provide implementation details of our spatiotemporal models, details on the evaluation, and the matching-based objective.
\subsection{Implementation Details}
We adopt ConditionalDETR~\cite{meng2021conditional} for our experiments due to its fast training. We use a ResNet50~\cite{he2016deep} backbone, pre-trained on ImageNet~\cite{deng2009imagenet}, and the same encoder and decoder hyperparameters as in ConditionalDETR. We train the approaches for future object detection using the same set-prediction loss function~\cite{meng2021conditional}. As a trade-off between performance and model training time, we set $T=2$ for our spatiotemporal approaches. We optimize with AdamW~\cite{loshchilov2018fixing}, using a batch size of 16 and weight decay of $10^{-4}$. We use base learning rates of $10^{-5}$ for the backbone and $10^{-4}$ for the encoder, decoder, and linear heads. The learning rates are warmed up for the first 10\% epochs and decayed by total factors of $\{0.5, 0.1\}$ when 60\% and 90\% of epochs have been processed. During the first 60\% epochs, we train with half resolution and double batch size. We train for 400 epochs on NuImages and 160 epochs on the relatively larger NuScenes.

\subsection{Evaluation Details}
We aim to predict objects present in a future frame. Thus, the detectors are fed images and expected to produce detections for a future time-step. Performance is measured in terms of Average Precision (AP). We report both average precision with a $0.5$ IoU threshold (AP50) and averaged over the thresholds $\{0.50,0.55,\dots,0.95\}$ (AP). We report both an average score over \emph{all} NuScenes classes and for the \texttt{car} and \texttt{pedestrian} classes. We also report results for objects of different sizes, using the cutoffs $H_0/24\cdot W_0/64$ and $H_0/4\cdot W_0/12$, where $(H_0, W_0)$ is the original image resolution.

\subsection{Matching-based Objective}
In DETR, the neural network is trained end-to-end for the object detection task by first matching the predicted objects, $\mathcal{O}$, to annotated objects, $\mathcal{A}$, and then, based on the matching, compute a loss. For future object detection, we adopt the same loss, but now computed against the annotated objects at time $T+\tau$, $\mathcal{A}^{T+\tau} = \{A_n^{T+\tau}\}_n$, where $n$ enumerates the annotated objects. To compute the matching, a matching score is computed between each pair of predicted and annotated objects,
\begin{align}
    \mathcal{L}^\text{match}(A_n^{T+\tau}, O_m^{T+\tau}) =& \mathcal{L}^\text{match}_\text{class}(A_n^{T+\tau}, \mathbf{c}_m^{T+\tau}) \\ &+ \mathcal{L}^\text{match}_\text{box}(A_n^{T+\tau}, \mathbf{b}_m^{T+\tau})\nonumber\enspace.
\end{align}
The class and bounding box of each predicted object $O_m^{T+\tau}$ is compared to each $A_n^{T+\tau}\in\mathcal{A}^{T+\tau}$. The matching scores are fed into the Hungarian algorithm that then computes the matching. Each predicted object $O_m^{T+\tau}$ is assigned an index $\sigma(m)$, corresponding to a matched annotated object or to background. The objective function is then computed as
\begin{align}
    \mathcal{L}(\mathcal{A}^{T+\tau}, \mathcal{O}^{T+\tau}) = [ & \sum_{m} \mathcal{L}_\text{class}(A_{\sigma(m)}^{T+\tau}, \mathbf{c}_m^{T+\tau}) \\&+ \mathbbm{1}_{\sigma(m)\neq\varnothing}\mathcal{L}_\text{box}(A_{\sigma(m)}^{T+\tau},\mathbf{b}_m^{T+\tau}) ] \nonumber\enspace.
\end{align}
Here, $\varnothing$ corresponds to \emph{no object}. We use the same matching cost functions $\mathcal{L}^\text{match}_\text{class}$ and $\mathcal{L}^\text{match}_\text{box}$, as well as the objective loss functions $\mathcal{L}_\text{class}$ and $\mathcal{L}_\text{box}$, as in Conditional DETR~\cite{meng2021conditional}.

\begin{figure*}[t]
    \centering
    \setlength{\tabcolsep}{0pt}
    \begin{tabular}{|c|c|}
        \hline
        \includegraphics[width=.49\linewidth, trim=0mm 1mm 0mm 0mm, clip]{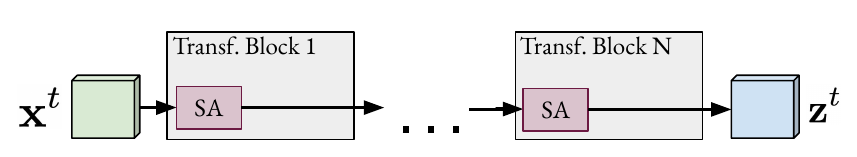} & \includegraphics[width=.49\linewidth, trim=0mm 1mm 0mm 0mm, clip]{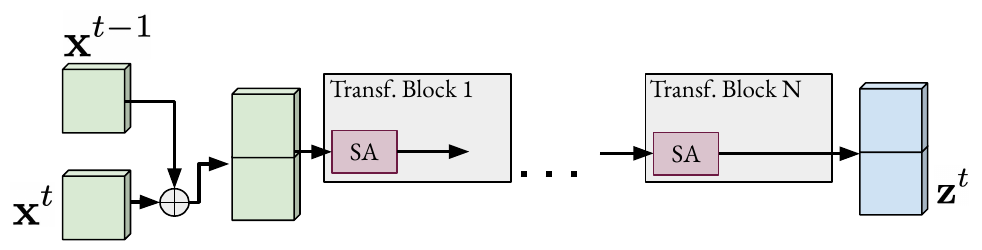}\\
        \hline
        \includegraphics[width=.49\linewidth, trim=0mm 1mm 0mm 0mm, clip]{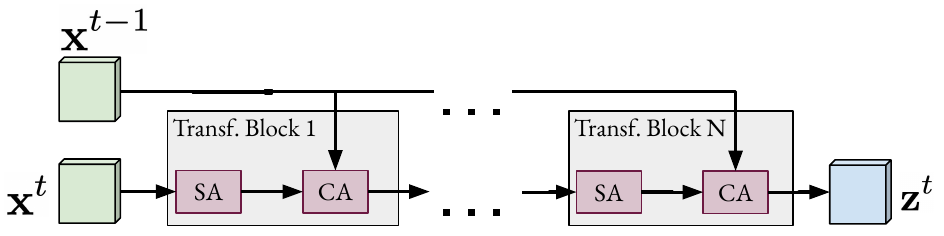} & \includegraphics[width=.49\linewidth, trim=0mm 1mm 0mm 0mm, clip]{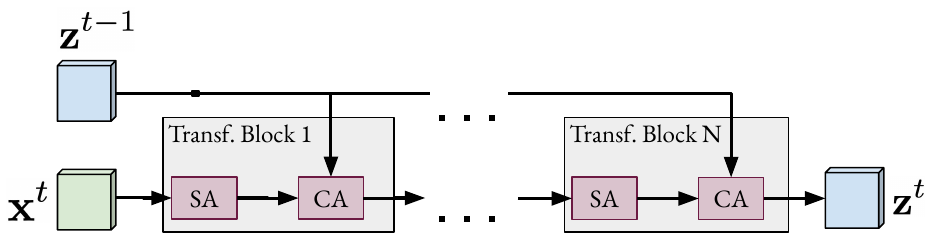}\\
        \hline
    \end{tabular}
    \caption{Illustration of the single-frame transformer encoder (top left) and three different spatiotemporal transformers: joint attention (top right), sequential cross-attention (bottom left), and recurrent transformer (bottom right).}
    \label{fig:encoders}
    \vspace{-0mm}
\end{figure*}

\begin{figure*}[t]
    \centering
    \setlength{\tabcolsep}{0pt}
    \begin{tabular}{|c|c|}
        \hline
        \includegraphics[width=.49\linewidth, trim=0mm 1mm 0mm 0mm, clip]{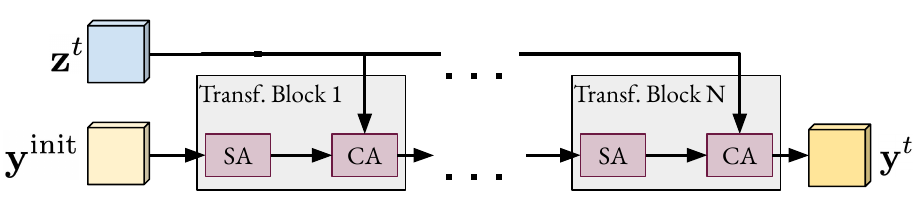} & \includegraphics[width=.49\linewidth, trim=0mm 1mm 0mm 0mm, clip]{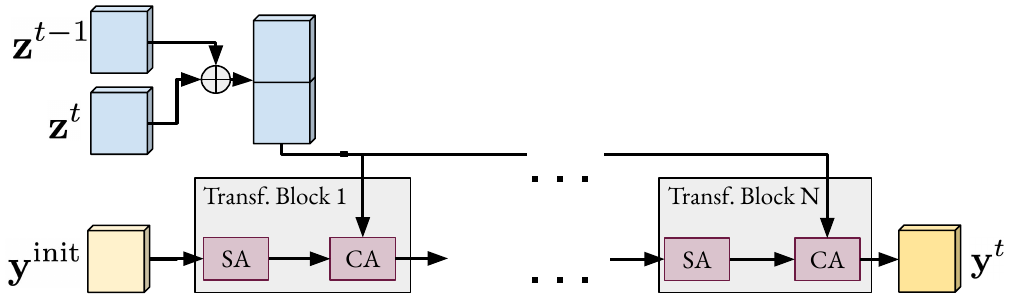}\\
        \hline
        \includegraphics[width=.49\linewidth, trim=0mm 0mm 0mm 0mm, clip]{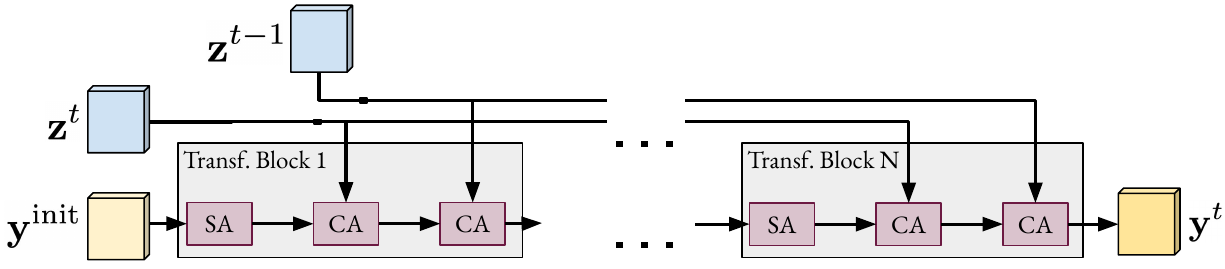} & \includegraphics[width=.49\linewidth, trim=0mm 2mm 0mm 0mm, clip]{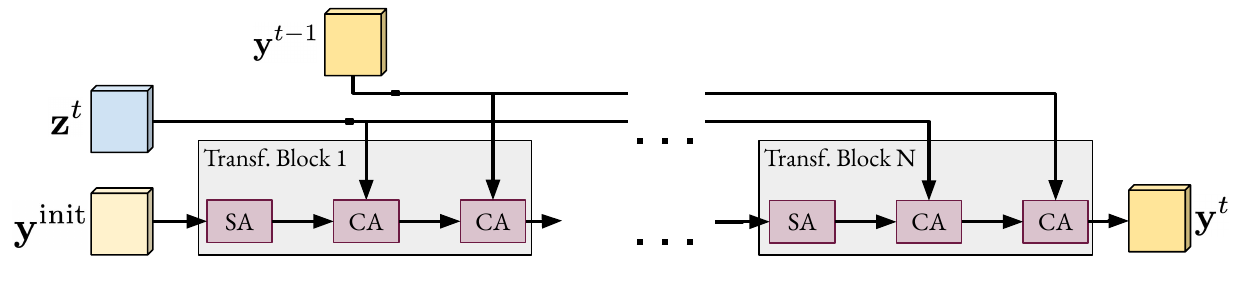}\\
        \hline
    \end{tabular}
    \caption{Illustration of the single-frame transformer decoder (top left) and three different spatiotemporal transformers: joint attention (top right), sequential cross-attention (bottom left), and recurrent transformer (bottom right).}
    \label{fig:decoders}
    \vspace{-0mm}
\end{figure*}

\subsection{Joint Attention and Recurrent Transformer}\label{appendix:spatiotemporal-mechanism}
In Section~\ref{sec:spatiotemporal}, three spatiotemporal mechanisms were mentioned, with an experimental comparison between them in Table~\ref{tab:ablations}. Only one of these mechanisms, the \emph{sequential cross-attention}, was described in detail in the main paper. Here, we describe the other two: the \emph{joint attention} and the \emph{recurrent transformer}.

\parsection{Joint Attention} One approach to processing spatiotemporal data is to jointly attend input from different timesteps. If done in the transformer encoder, this would take the form
\begin{align}
    \mathbf{z}^T = (\enc([\mathbf{x}^1,\dots,\mathbf{x}^T]))_T\enspace.
\end{align}
Here, $[\cdot,\cdot]$ denotes stacking the feature maps in a new, temporal dimension. Within the transformer encoder, the temporal dimension is flattened together with the two spatial dimensions. The joint attention operation is applied over all feature vectors $\{\mathbf{x}_{i,j}^t\in\mathbb{R}^D: i=1,\dots,H,\, j=1,\dots,W,\,t=1,\dots,T\}$. After applying the transformer encoder, only the feature vectors corresponding to the time $t=T$ are picked out, which is denoted by $(\cdot)_T$. It should be noted that each feature vector $\mathbf{x}_{i,j}^t$ attends all feature vectors, leading to a quadratic complexity in $T$.

If the joint attention mechanism is instead introduced in the decoder, we get
\begin{align}
    \mathbf{y}^T = \dec(\mathbf{y}^\text{init}, [\mathbf{z}^1,\dots,\mathbf{z}^T])\enspace.
\end{align}
The queries $\mathbf{y}^\text{init}$ jointly cross-attend all features vectors. As the number of queries is constant regardless of $T$, the complexity becomes linear in $T$.

A crucial component of the attention operation is the positional encoding. For the joint attention, we augment the positional encoding used in DETR~\cite{carion2020end} and ConditionalDETR~\cite{meng2021conditional} by additionally encoding the temporal position, in seconds, to the spatial position. This enables the joint attention mechanism to take the time at which a feature vector was captured into account.

\parsection{Recurrent Transformer} Recurrent neural networks enables processing time series. We use the attention mechanism to attend the encoder or decoder output in the previous frame. The resulting neural network \emph{is} a recurrent neural network. In principle, this allows the network to learn an explicit tracking solution, where detections in one frame are used as the starting point in the next frame. This mechanism closely follows the sequential cross-attention mechanism. The difference is that instead of attending input from previous frames, it is the output that is observed. For the encoder, this takes the following form. A self-attention layer,
\begin{align}
    \mathbf{z}^{t,i,0} = \enc^{i,0}(\mathbf{z}^{t,i-1}, \mathbf{z}^{t,i-1})\enspace,
    \label{eq:rca-enc-1}
\end{align}
is followed by a recurrent cross-attention layer,
\begin{align}
    \mathbf{z}^{t,i} = \enc^{i,1}(\mathbf{z}^{t,i,0}, \mathbf{z}^{t-1,L})\enspace.
    \label{eq:rca-enc-2}
\end{align}
For $t=1$, we skip the recurrent cross-attention layer and we let $\mathbf{z}^{t,0}=\mathbf{z}^t$. Note that computation of $\mathbf{z}^{t}$ in \eqref{eq:rca-enc-1}-\eqref{eq:rca-enc-2} relies on the existence of $\mathbf{z}^{t-1}$.

If the recurrent cross-attention mechanism is applied in the decoder, we get
\begin{align}
    \mathbf{y}^{t,i,0} = \dec^{i,0}(\mathbf{z}^{t,i-1}, \mathbf{y}^{t,i-1})\enspace,
\end{align}
followed by
\begin{align}
    \mathbf{y}^{t,i} = \dec^{i,1}(\mathbf{y}^{t,i,0}, \mathbf{y}^{t-1,L})\enspace.
\end{align}
As with the decoder, the recurrent cross-attention layer is skipped for $t=1$. We assign $\mathbf{y}^{t,0}=\mathbf{y}^\text{init}$. Again, note that computation of $\mathbf{y}^t$ relies on the existence of $\mathbf{y}^{t-1}$.

\section{Architecture Illustrations}
We provide illustrations for the detection transformer baseline, the joint attention, the sequential cross-attention, and the recurrent transformer. We show the encoder in Figure~\ref{fig:encoders} and the decoder in Figure~\ref{fig:decoders}

\section{Additional Experimental Results}\label{appendix:additional-experimental}
We provide detailed tables for the ablation experiments, an analysis of $\tau > 1$, and additional qualitative evaluations.

\subsection{Detailed Ablation Results}

The ablation results in the main manuscript were limited to the performance on the car and pedestrian class, omitting the class average as well as size breakdowns. We provide more extensive results on the spatiotemporal ablation in Table~\ref{tab:appendix_architecture}. 

We also performed an ablation study of the ego-motion mechanism. In Section-\ref{sec:egomotion} we described how to encode the ego-motion information and incorporate it into the network in a learnable manner, either in the encoder or the decoder. We evaluate the effectiveness of these approaches in Table~\ref{tab:appendix_egomotion}. As already established, ego-motion is a powerful cue that leads to significantly improved performance. However, the specific location where ego-motion is incorporated seems of minor importance. Here we choose to add it in the encoder.

\begin{table*}[t]
    \centering
    \resizebox{\linewidth}{!}{%
    \begin{tabular}{l c c c c c c c c c}
        \toprule
        \multirow{2}{*}{\textbf{Method}} & \multicolumn{3}{c}{\textbf{AP50}} & \multicolumn{3}{c}{\textbf{AP50}} & \multicolumn{3}{c}{\textbf{AP}}\\
         & Mean & Car & Pedestrian & Small & Medium & Large & Mean & Car & Pedestrian\\
        \midrule
        Singleframe                 & 13.5 & 22.5 &  6.2 &  3.0 & 12.9 & 30.8 & 3.8 &  6.7 &  1.5\\
        Joint Attention Encoder     & 22.9 & 34.8 & 10.5 &  3.8 & 23.1 & 49.2 & 7.4 & 11.8 &  2.7\\
        Joint Attention Decoder     & \second{23.5} & \second{35.4} & \second{11.0} & \second{4.2} &  \second{23.7} & \second{52.0} &  \second{7.9} & \second{11.9} &  \second{2.8}\\
        Sequential CA Encoder       & 21.9 & 35.0 &  9.7 &  \first{4.7} &  21.9 & 48.1 &  7.1 & \second{11.9} &  2.5\\
        Sequential CA Decoder       & \first{23.9} & \first{36.6} & \first{12.0} & 4.1 & \first{25.1} & \first{52.1} &  \first{8.1} & \first{12.1} &  \first{3.1}\\
        Recurrent Tr. Encoder        & 21.6 & 34.9 &  9.5 &  3.7 &  21.9 & 44.2 &  7.3 & 11.8 &  2.4\\
        Recurrent Tr. Decoder        & 21.2 & 32.6 &  9.2 &  3.6 &  22.1 & 42.4 &  6.6 & 10.8 &  2.3\\
        \bottomrule
    \end{tabular}
    }
    \caption{Future object detection performance for different spatiotemporal mechanisms on the NuImages validation set. The best mechanism is marked in \first{red} and the second best in \second{blue}.}
    \label{tab:appendix_architecture}
    \vspace{-5mm}
\end{table*}

\begin{table*}[t]
    \centering
    \resizebox{\linewidth}{!}{%
    \begin{tabular}{l c c c c c c c c c}
        \toprule
        \multirow{2}{*}{\textbf{Method}} & \multicolumn{3}{c}{\textbf{AP50}} & \multicolumn{3}{c}{\textbf{AP50}} & \multicolumn{3}{c}{\textbf{AP}}\\
         & Mean & Car & Pedestrian & Small & Medium & Large & Mean & Car & Pedestrian\\
        \midrule
        No ego-motion                & 23.9 & 36.6 & 12.0 & 4.1 & 25.1 & 52.1 & 8.1 & 12.1 &  3.1 \\
        Ego-motion in encoder & 28.0 & \first{43.2} & \first{15.1} & \second{4.7} & \second{29.9} & \first{59.3} & 9.4 & \first{15.5} & 3.4 \\
        Ego-motion in decoder & \second{28.3} & \second{42.4} & 13.7 & \second{4.7} & 29.8 & \second{59.1} & \second{10.0} & \second{14.9} & \second{3.6}\\
        \bottomrule
    \end{tabular}
    }
    \caption{Future object detection performance on the NuImages validation set with and without ego-motion.}
    \label{tab:appendix_egomotion}
    \vspace{-0mm}
\end{table*}

\subsection{Analysis for $\tau > 1$}

 In the main manuscript all experiments use the same input time gap as the prediction horizon, $\tau=1$. Meaning that if the method predicts 500 ms into the future, it observes two frames that are also 500 ms apart. Here we add additional experiments to show that this is not an inherent limitation of our method. In Table~\ref{tab:pred_horizon}, we trained networks to observe frames 50 ms apart and make a prediction at $\tau=\{1, 2, 5, 10\}$, corresponding to 50 ms, 100 ms, 250 ms, 500 ms. In this setting, our approach still substantially outperforms the baselines in Table~4 of the main paper, and performs similarly to when the input gap is equal to the prediction horizon (bottom row).

\begin{table*}[t]
    \centering
    \resizebox{0.7\linewidth}{!}{%
    \begin{tabular}{l c c c c c c c c}
        \toprule
        \textbf{Input} & \multicolumn{2}{c}{\textbf{50 ms}} & \multicolumn{2}{c}{\textbf{100 ms}} & \multicolumn{2}{c}{\textbf{250 ms}} & \multicolumn{2}{c}{\textbf{500 ms}}\\
        \textbf{spacing} & Car & Pedestrian & Car & Pedestrian & Car & Pedestrian & Car & Pedestrian\\
        \midrule
        50 ms     & 72.4 & 49.4 & 69.5          & \first{44.2} & \first{66.1} & 35.2 & 52.5 & 19.7\\
        as horizon & 72.4 & 49.4          & \first{70.1} & \first{44.2} & 65.6 & \first{35.8} & \first{54.9} & \first{21.5}\\
                \bottomrule
    \end{tabular}
    }
    \caption{Different input spacing strategies on NuScenes validation set (compare to Table~4 of main paper). Note that for 50 ms, both rows refer to the same experiment.}
    \label{tab:pred_horizon}
    \vspace{-0mm}
\end{table*}